\documentclass{article}

\usepackage{microtype}
\usepackage{graphicx}
\usepackage{booktabs} 

\usepackage{hyperref}

\usepackage[accepted]{icml2023}

\usepackage{amsmath}
\usepackage{amssymb}
\usepackage{mathtools}
\usepackage{amsthm}

\usepackage{multicol}
\usepackage{multirow}
\usepackage{makecell}
\usepackage{subcaption}

\usepackage[capitalize,noabbrev]{cleveref}

\theoremstyle{plain}

\theoremstyle{definition}

\theoremstyle{remark}

\usepackage[textsize=tiny]{todonotes}

\icmltitlerunning{Conformal Prediction with Large Language Models for Multi-choice Question Answering}

\begin{document}

\twocolumn[
\icmltitle{Conformal Prediction with Large Language Models \\ for Multi-Choice Question Answering}



\icmlsetsymbol{equal}{*}

\begin{icmlauthorlist}
\icmlauthor{Bhawesh Kumar}{equal,yyy}
\icmlauthor{Charles Lu}{equal,xxx}
\icmlauthor{Gauri Gupta}{xxx}
\icmlauthor{Anil Palepu}{zzz}
\icmlauthor{David Bellamy}{yyy}
\icmlauthor{Ramesh Raskar}{xxx}
\icmlauthor{Andrew Beam}{yyy}
\end{icmlauthorlist}

\icmlaffiliation{yyy}{Harvard University}
\icmlaffiliation{xxx}{MIT Media Lab}
\icmlaffiliation{zzz}{Harvard-MIT Health Sciences \& Technology}

\icmlcorrespondingauthor{Bhawesh Kumar}{bhaweshk@mit.edu}
\icmlcorrespondingauthor{Charlie Lu}{luchar@mit.edu}

\icmlkeywords{Machine Learning, ICML}

\vskip 0.3in
]



\printAffiliationsAndNotice{\icmlEqualContribution} 

\begin{abstract}
As large language models are widely developed, robust uncertainty quantification techniques will become crucial for safe deployment in high-stakes scenarios. 
This work explores how conformal prediction can quantify uncertainty in language models for multiple-choice question-answering. 
We find that the uncertainty estimates from conformal prediction are tightly correlated with prediction accuracy. 
This observation can be helpful in downstream applications such as selective classification and filtering out low-quality predictions.
We also investigate the exchangeability assumption required by conformal prediction to out-of-subject questions, which may be a more realistic scenario for many practical applications. 
Our work contributes towards more trustworthy and reliable usage of large language models in safety-critical situations, where robust guarantees of error rate are required.
\end{abstract}

\section{Introduction} \label{intro}

Large language models (LLMs) have recently achieved impressive performance on a number of NLP tasks, such as machine translation, text summarization, and code generation. However, lingering concerns of trust and bias still limit their widespread application for critical decision-making domains such as healthcare.

One well-known issue with current LLMs is their tendency to ``hallucinate'' false information with seemingly high confidence. These hallucinations can occur when the model generates outputs not grounded in any factual basis or when the prompt is highly unusual or ambiguous. This behavior of LLMs may also result from how these models are trained --- using statistical sampling for next-token prediction --- which can progressively increase the likelihood of factual errors as the length of generated tokens increases~\cite{LeCun_2023}. Factually incorrect outputs may confuse and deceive users into drawing wrong conclusions, ultimately decreasing the overall system's trustworthiness. Decisions based on unpredictable or biased model behavior could have significant negative and socially harmful consequences in high-stakes domains such as healthcare and law.

Therefore, we seek to explore principled uncertainty quantification (UQ) techniques for LLMs that can provide guaranteed error rates of model predictions. Ideally, these UQ techniques should be model agnostic and easy to implement without requiring model retraining due to the intensive computing costs and limited API access associated with many LLMs. To this end, we investigate \textit{conformal prediction}, a distribution-free UQ framework, to provide LLMs for the task of multiple-choice question-answering (MCQA).  

Based on our experiments, we find the uncertainty, as provided by conformal prediction, to be strongly correlated with accuracy, enabling applications such as filtering out low-quality predictions to prevent a degraded user experience. We also verify the importance of the exchangeability assumption in conformal prediction (see section \ref{background}) for guaranteeing a user-specified level of errors.

To summarize, our contributions are the following:
\begin{itemize}
\item we adapt conformal prediction for MCQA tasks to provide distribution-free uncertainty quantification in LLMs,
\item show how the uncertainty provided by conformal prediction can be useful for downstream tasks such as selective classification,
\item and assess the performance of conformal prediction when the exchangeability assumption is violated for in-context learning in LLMs.
\end{itemize}

\vfill
\section{Conformal Prediction} \label{background}

Uncertainty quantification (UQ) techniques are critical to deploying machine learning in domains such as healthcare~\cite{bhatt2021uncertainty,kompa2021second, kompa2021empirical}. Conformal prediction~\cite{gammerman2013learning, Vovk_2022} is a flexible and statistically robust approach to uncertainty quantification. Informally, the central intuition behind conformal prediction is to output a set of predictions containing the correct output with a user-specified probability. 

By providing a more nuanced understanding of the model's confidence and a statistically robust coverage guarantee, conformal prediction paves the way for improved and more reliable applications of machine learning models across various domains~\cite{kumar2022reliable}.

\textbf{Prediction sets.}
Formally, let $\mathcal{C}: \mathcal{X} \rightarrow 2^\mathcal{Y}$ be a set-valued function that generates a prediction sets over the powerset of $Y$ given an input $X$. 
This prediction set naturally encodes the model's uncertainty about any particular input by the \textbf{size} of the prediction set. 

Expressing uncertainty as the set size is an intuitive output that can be helpful in decision-making contexts~\cite{babbar2022utility}. For example, in medical diagnosis, the concept of prediction set is similar to a differential diagnosis, where only likely and plausible conditions are considered given the observed symptoms of a patient~\cite{lu2022fair}. Indeed, conformal prediction has been utilized for uncertainty quantification in healthcare applications such as medical imaging analysis~\cite{lu2022improving,lu2022three,lu2022distributionfree}. 

\textbf{Coverage guarantee.}
Conformal methods generate prediction sets that ensure a certain user-specified probability of containing the actual label, regardless of the underlying model or distribution. This guarantee is achieved without direct access or modification to the model's training process and only requires a held-out calibration and inference dataset.
This makes conformal prediction well-suited to LLM applications when retraining is costly and direct model access is unavailable through third-party or commercial APIs.

The coverage guarantee states that the prediction sets obtained by conformal prediction should contain the true answer on average at a user-specified \textit{level}, \emph{$\alpha$}. This property is called \textit{coverage}, and the corresponding coverage guarantee is defined as:
\begin{equation}
    \label{eq:coverage-guarantee}
    1 - \alpha \leq \mathbf{P}\left(Y_\text{test} \in \, \mathcal{C}(X_\text{test})\right),
\end{equation}
where $\alpha \in (0, 1)$ is the desired error rate, and $\mathcal{C}$ is the calibrated prediction set introduced above. ${\left(X_\text{test}, Y_\text{test}\right) \sim \mathcal{D_\text{calibration}}}$ is an unseen test point that is drawn from the same distribution as the data used to calibrate the prediction sets.

    
 
\textbf{Conformal Calibration Procedure.}
    As previously mentioned, conformal prediction only needs the scores of a model to calibrate and construct the prediction sets. We now describe how to calibrate the prediction sets for a specific score function.
    
    Let $f: \mathcal{X} \rightarrow \Delta^{\lvert \mathcal{Y} \rvert}$ be a classifier with a softmax score, where $\Delta$ is a $\lvert\mathcal{Y}\rvert$-dimensional probability simplex.
    A common choice for the score function, \textit{least ambiguous set-valued classifiers} (LAC)~\cite{doi:10.1080/01621459.2017.1395341}, is defined as
    \begin{equation}
        \label{eq:score-function}
        S(X, Y) = 1 - \left[f(X)\right]_{Y}, 
    \end{equation}
    where $\left[f(X)\right]_Y$ is the softmax score at the index of the true class.
    
    To calibrate the prediction sets to our desired level of coverage, we need to estimate a threshold $\hat{q}_\alpha$ that is the $1-\alpha$ quantile of the calibration scores 
    \begin{equation}
    \hat{q}_\alpha = \text{Quantile}\left(\{s_1, \ldots, s_n\}, \frac{\lceil{(n+1)(1-\alpha)\rceil}}{n}\right),
    \end{equation}
    where $\{s_1, \ldots, s_n\}$ are the LAC scores of the calibration set.
    
    At inference time, prediction sets can be constructed in the following manner:
    \begin{equation}
        \label{eq:prediction-set}
        \mathcal{C}(X) = \left\{y \in \mathcal{Y}: S(X, y) \leq \hat{q}_\alpha \right\},
    \end{equation}

\textbf{Exchangeability assumption.}
Conformal prediction assumes that the data used to calibrate the prediction sets is exchangeable with the test data at inference time.
If this assumption holds, the coverage guarantee, as stated in Equation~\ref{eq:coverage-guarantee}, will hold, and the resulting prediction sets will have the desired error rate.

Exchangeability can be viewed as weaker than the independent and identically distributed (IID) assumption~\cite{bernardo1996concept}. This assumption is often made in machine learning with regard to the training, validation, and test sets.
The threshold used to determine the size of the prediction set is estimated on a held-out calibration data set that is assumed to be \textit{exchangeable} with the test distribution.

\begin{figure}[ht]
\vskip 0.2in
\begin{center}
    \includegraphics[width=\columnwidth]{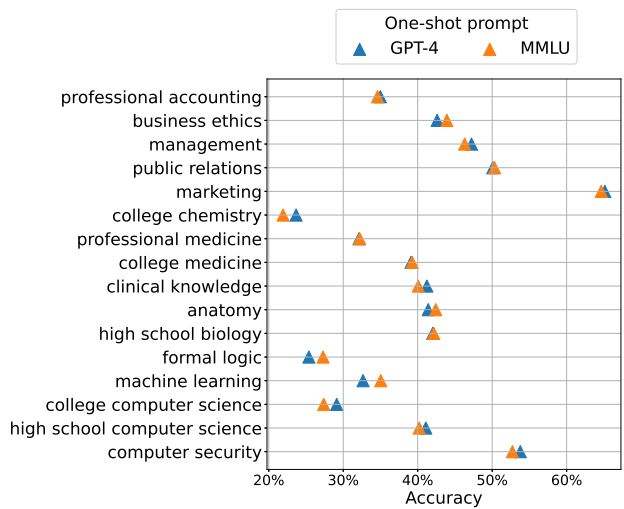}
    \vspace{-0.1in}
    \caption{\textbf{LLaMA MCQA accuracy is similar for GPT-4 generated questions and real MMLU questions across subjects.} For most MMLU subjects, prediction accuracy using one-shot GPT-4 generated questions is similar to when actual MMLU questions are used in one-shot prompts. Results are averaged over ten randomly selected one-shot GPT-4 and MMLU prompts.}
    \label{fig:prompt}
\label{fig:gpt-vs-mmlu}
\end{center}
\vskip -0.2in
\end{figure}

\begin{figure}[ht]
\vskip 0.2in
\begin{center}
    \includegraphics[width=\columnwidth]{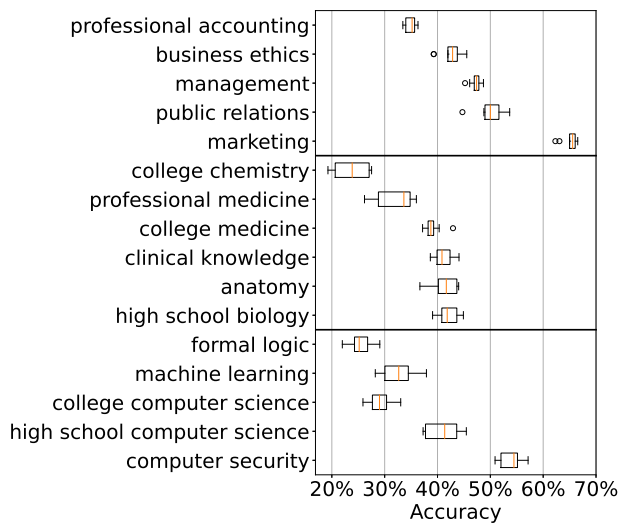}
    \caption{\textbf{The accuracy distribution across subjects for ten prompts.} We plot the distribution of accuracy for ten different one-shot prompts.}
\label{fig:accuracy}
\end{center}
\vskip -0.2in
\end{figure}

\section{Prompt Engineering} \label{approach}
In this paper, we focus on the task of multiple-choice question answering (MCQA) and frame MCQA as a supervised classification task, where the objective is to predict the correct answer choice out of four possible options.
We wish to quantify the model uncertainty over the predicted output using conformal prediction. 
We condition each option choice (A, B, C, and D) on the prompt and question and use the LLaMA-13B model~\cite{touvron2023llama} to generate the logit corresponding to each multiple-choice answer. We normalize the four logits using the softmax to obtain valid probabilities for each option.

\textbf{One-shot prompting.} \label{one-shot}
LLMs are very sensitive to the exact input prompt, which has motivated a whole field of in-context learning and prompt engineering or prompt tuning~\cite{zhou2023leasttomost,wei2023chainofthought}. Context learning refers to the ability of LLMs to understand and make predictions based on the context in which the input data is presented without updating the model weights. Prompt engineering methods vary significantly among tasks and require heavy experimentation and reliance on hand-crafted heuristics. For the current setup, model performance on classification tasks is often sensitive to the prompts used. Thus, we experiment with several prompting strategies before finalizing our prompts.

We use one-shot prompting by including one context example. For each subject, we use a slightly different prompt. For example, we prompt the model to assume it is the ``world's best expert in college chemistry'' when generating predictions for college chemistry subjects. 

We also use ten different prompts for each subject to generate ten softmax probability outputs to reduce variance. We obtain the final probability outputs for a question by averaging the softmax outputs corresponding to these ten prompts. The ten prompts for a given subject only vary in terms of the one-shot question. A sample prompt for high school biology is provided below:

\begin{verbatim}
This is a question from high school 
biology. 

A piece of potato is dropped into a 
beaker of pure water. Which of the 
following describes the activity after 
the potato is immersed into the water?
(A) Water moves from the potato into 
the surrounding water. 
(B) Water moves from the surrounding 
water into the potato. 
(C) Potato cells plasmolyze. 
(D) Solutes in the water move into 
the potato. 
The correct answer is option B. 

You are the world's best expert in 
high school biology. Reason 
step-by-step and answer the 
following question. 
From the solubility rules, which of 
the following is true? 
(A) All chlorides, bromides, and 
iodides are soluble 
(B) All sulfates are soluble 
(C) All hydroxides are soluble 
(D) All ammonium-containing compounds 
are soluble 

The correct answer is option:
\end{verbatim}
\textbf{GPT-4 generated examples.}
 We explore two approaches for the one-shot example in the prompts: (1) One-shot example is one of the questions in the MMLU dataset for that subject. We then exclude this specific question for generating predictions with the resulting prompt. (2) We use GPT-4 to generate multiple-choice questions for each subject. We then cross-check the questions and answers produced by GPT-4 for correctness and select ten correct question-answer pairs. 
 
 We use the following prompt to generate MCQs for clinical knowledge from GPT-4: ``\textit{Give me 15  multiple choice questions on clinical knowledge with answers}''. Specific questions and answers generated by the GPT-4 are available from our code (refer to Section~\ref{code}.) We have also included a subset of sample GPT-4 generated questions and answers as well as MMLU-based questions and answers in the Appendix~(\ref{appendix_qa} )

We generate MCQs for other subjects using similar prompts.
GPT-4-based one-shot questions produce more accurate answers than MMLU-based questions, as shown in Figure~\ref{fig:gpt-vs-mmlu}. 

After controlling for the size of the prompts (limited to 700 tokens), we find that MMLU-based and GPT-4 based one-shot questions produce similar accuracy on the sixteen subjects we evaluate. We conduct all the following experiments on prompts that use GPT-4-based one-shot questions since they are shorter on average and achieve similar performance.

\begin{figure}[ht]
\vskip 0.2in
\begin{center}
    \includegraphics[width=\columnwidth]{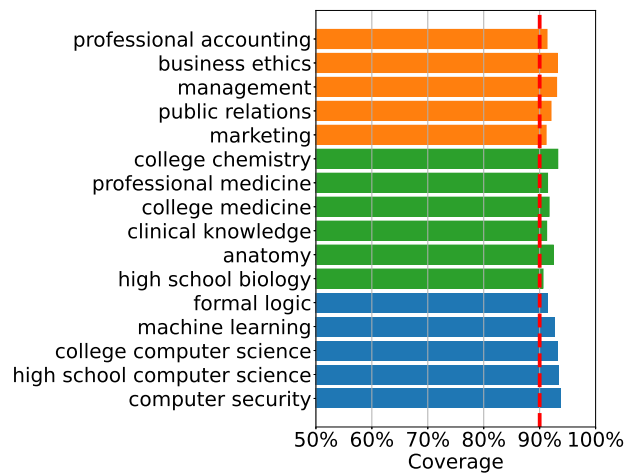}
    \caption{\textbf{Desired coverage is achieved for all subjects.} The red dashed line shows the desired coverage rate (specified at $\alpha=0.1$), which is guaranteed by conformal prediction to be with at least $1-\alpha$ percent of the time. The colors denote the three categories of questions.}
\label{fig:coverage}
\end{center}
\vskip -0.2in
\end{figure}

\begin{figure}[ht]
\vskip 0.2in
\begin{center}
    \includegraphics[width=\columnwidth]{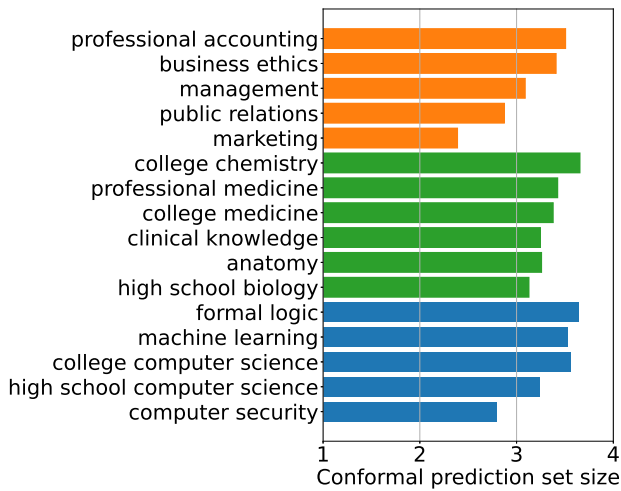}
    \caption{\textbf{Uncertainty quantification using prediction set size.} In conformal prediction,  a set of predictions is generated for each question. The size of this set indicates how uncertain the model is for a particular question. Larger set sizes denote greater uncertainty, and smaller set sizes denote less uncertainty. The colors denote the three categories of questions.}
\label{fig:set-size}
\end{center}
\vskip -0.2in
\end{figure}

\begin{figure}[ht]
\vskip 0.2in
\begin{center}
    \includegraphics[width=\columnwidth]{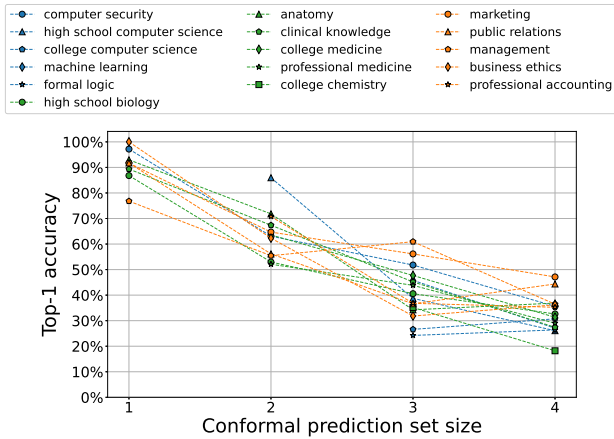}
    \caption{\textbf{Top-1 accuracy stratified by prediction set size.} For all subjects, we find a strong correlation between the prediction uncertainty (as measured by set size) and the top-1 accuracy of those predictions. Conformal prediction can be used for selective classification by filtering those predictions in which the model is highly uncertain.}
\label{fig:selective-classification}
\end{center}
\vskip -0.2in
\end{figure}

\begin{figure}[ht]
\vskip 0.2in
\begin{center}
    \includegraphics[width=\columnwidth]{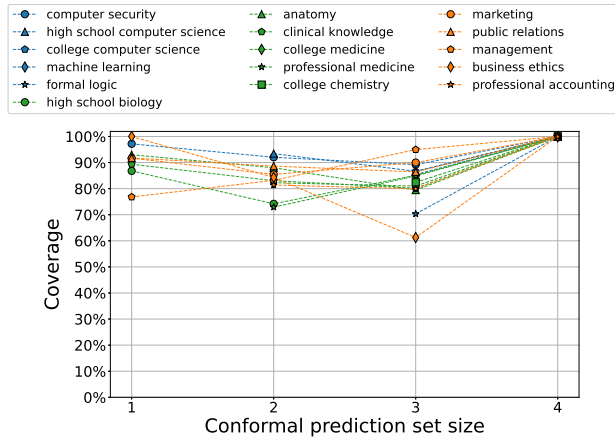}
    \caption{\textbf{Stratified coverage at each size of prediction set.} For most subjects, coverage is fairly consistent at all set sizes for prediction sets constructed with the conformal prediction procedure at $\alpha = 0.1$. This means that the true answer is one of the items in the predicted set on average about $90\%$ of the time.}
\label{fig:stratified-coverage}
\end{center}
\vskip -0.2in
\end{figure}

\begin{figure}[ht]
\vskip 0.2in
\begin{center}
    \includegraphics[width=\columnwidth]{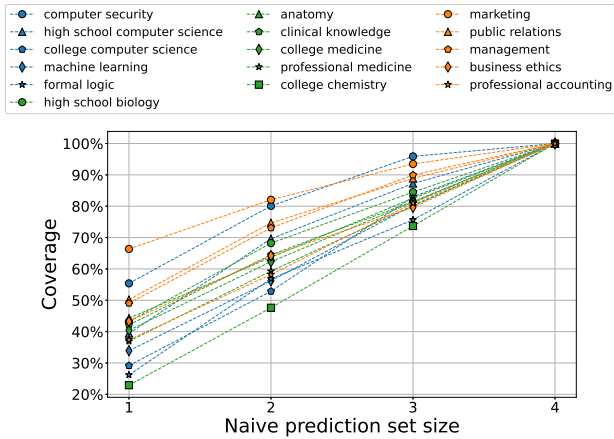}
    \caption{\textbf{Coverage of naive top-$k$ prediction sets.} Coverage sharply falls off at smaller set sizes for naive prediction sets constructed by simply taking the top-$k$ softmax scores for all predictions.}
\label{fig:naive-sets}
\end{center}
\vskip -0.2in
\end{figure}

\begin{figure*}[ht]
\vskip 0.2in
\begin{center}
    \includegraphics[width=0.9\textwidth]{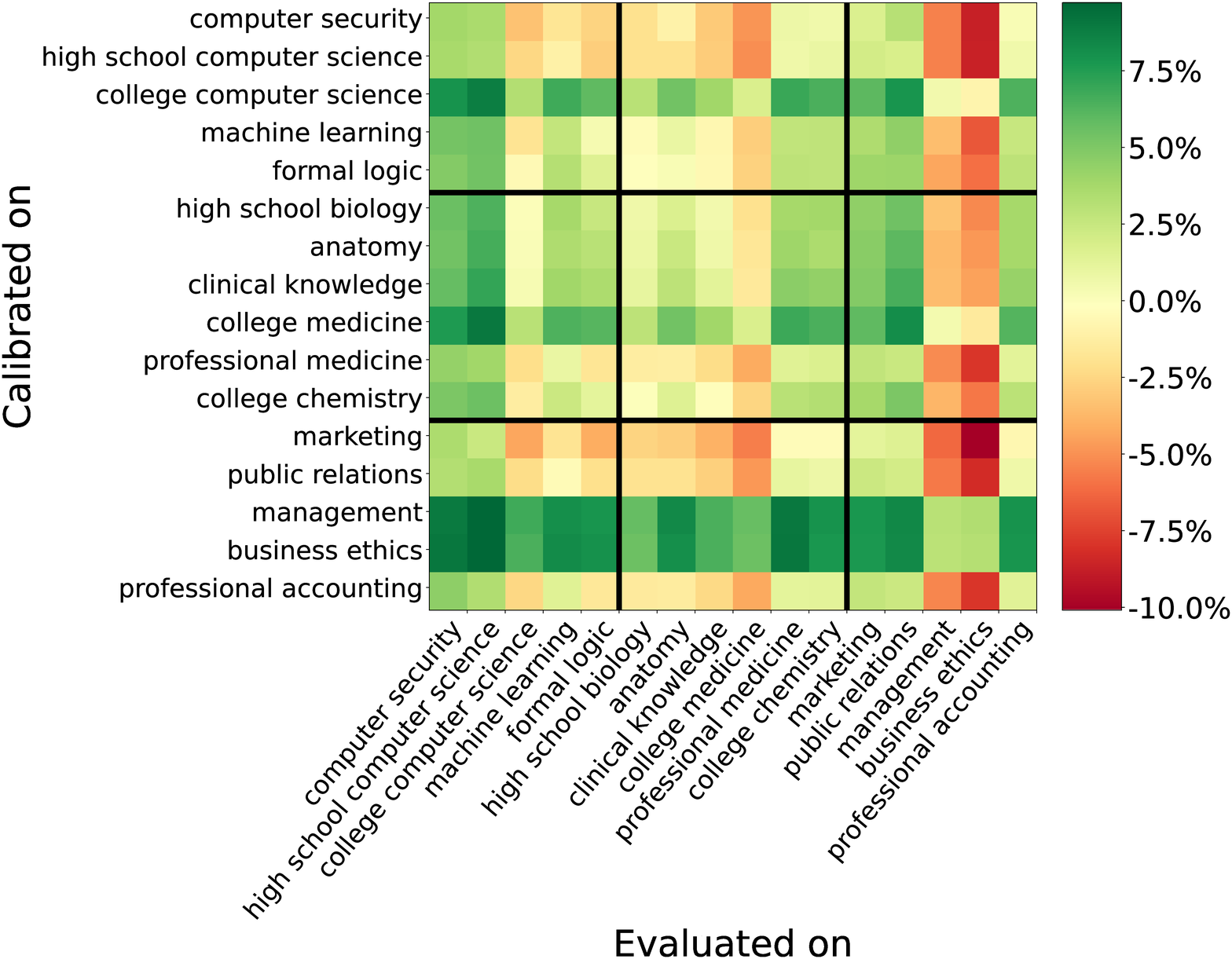}
    \caption{\textbf{Difference in coverage when calibrated on different subjects.} Deviation from $90\%$ coverage for $\alpha=0.1$. The off-diagonals represent entries corresponding to the cases where exchangeability conditions are violated between calibration and evaluation data sets. The subjects are grouped into the three broad categories of computer science, medicine, and business.}
\label{fig:cross-calibration}
\end{center}
\vskip -0.2in
\end{figure*}

\section{Experiments}
\subsection{Model and dataset} \label{model_and_dataset}
    \label{sub:datasets}
    We use the LLaMA-13B model~\cite{touvron2023llama} to generate predictions for MCQA. LLaMA-13B is an open-source 13 billion parameter model trained on 1 trillion tokens and has been shown to achieve good zero-shot performance on various question-answering benchmarks. For our dataset, we use the MMLU benchmark~\cite{hendrycks2021measuring}, which contains MCQA questions from 57 domains covering subjects such as STEM, humanities, and medicine. 

    For our experiments, we considered the following subset of MMLU: computer security, high school computer science, college computer science, machine learning, formal logic, high school biology, anatomy, clinical knowledge, college medicine, professional medicine, college chemistry, marketing, public relations, management, business ethics, and professional accounting. We group these domains into three broad categories: ``business'', ``medicine'', and ``computer science''. These 16 subjects represent diverse domains and have sufficient samples (each with at least 100 questions). 

    We perform classification by obtaining logit scores corresponding to option choices `A', `B', `C', and `D' conditioned on the one-shot prompt and the question. For example, for the sample prompt and question pair described in section \ref{one-shot}, we find the logit score corresponding to the next tokens corresponding to each of the four options. We then take the softmax over the logit scores corresponding to the options choices to obtain probability scores. The softmax scores corresponding to ten different prompts (that vary in terms of one-shot questions) are averaged to obtain final probability scores for each question-option pair.
    
\subsection{Setup} 
    \label{sub:setup}
    We randomly split the data into equal-sized calibration and evaluation sets for each subject and averaged results over 100 random trials for our conformal prediction experiments. 
    For each trial, we randomly sample $50\%$ of data for calibration and $50\%$ to evaluate coverage and set size. Thus, we have at least 50 samples for calibration. While the theoretical guarantee of conformal prediction holds on average for even such a small number of calibration samples, the individual 100 random trials may not always have exact coverage. A higher calibration size can reduce variance in coverage associated with the different random trials~\cite{gentle22}.

\subsection{Results} \label{results}
\textbf{Naive Calibration in LLMs.} 
    Previous works have studied calibration in context to LLMs. \cite{si-etal-2022-examining} looked at the limitation of traditional calibration metrics like Expected Calibration Error (ECE) and Maximum Calibration Error (MCE). \cite{jiang-etal-2021-know} looked at T5, BART, and GPT-2 language models and found that the models are not calibrated for question-answering tasks. More recently, \cite{kadavath2022language} found that large language models are well calibrated for various MCQA tasks. 
    In current work, we examine the calibration error in the softmax probability output for the MCQA task for the LLaMA-13B language model. To this end, we calculate the Expected Calibration Error (ECE) and Maximum Calibration Error (MCE), metrics that measure the average and maximum discrepancy between the confidence of the model's predictions and their accuracy. We find that the naive softmax output of the model is reasonably well calibrated across subjects on average, with ECE varying between a minimum of 1\% for high school biology to a maximum of 7\% for marketing (refer figure \ref{fig:calibration-error} in the appendix.) This aligns with previous findings on calibration error in LLMs~\cite{kadavath2022language}. Nonetheless, MCE is significant for most subjects, indicating that the model is under-confident or over-confident at specific confidence levels. Additionally, there are no formal guarantees in terms of calibration errors.

\textbf{Difference in coverage and set sizes between subjects.}
    We next implement the conformal prediction procedure and compare coverage and prediction set size between subjects in Figure~\ref{fig:coverage} and Figure~\ref{fig:set-size} at the error rate $\alpha = 0.1$. 
    The coverage guarantee of conformal prediction holds across all subjects (Figure~\ref{fig:coverage}). Comparing Figure~\ref{fig:accuracy} and Figure~\ref{fig:set-size}, we see that for each of the three categories, uncertainty --- as measured by prediction set sizes --- is, in general, significant for subjects with low top-1 accuracy and low for subjects with high top-1 accuracy. 
    
    For example, more challenging subjects such as formal logic and college chemistry have the most uncertainty on average, while ``easier'' subjects such as marketing have the lower average uncertainty.
    We show more results for different $\alpha$ values in Table~\ref{tab:results}.

\textbf{Selective classification with conformal prediction.}
    Conformal Prediction framework can also be used for selective classification~\cite{angelopoulos2022learn, angelopoulos2021gentle}. In Figure~\ref{fig:selective-classification}, we analyze the correlation between uncertainty (as measured by conformal prediction) and top-1 accuracy performance. Specifically, we look at top-1 accuracy across subjects stratified by the size of the prediction set outputted by conformal prediction. We find a robust negative correlation between set size and top-1 accuracy for all subjects. This is intuitive as models with low confidence scores should correspond to less accurate predictions. 
    
    The accuracy for prediction sets with only one prediction is significantly higher than naive top-1 accuracy, as shown in Figure~\ref{fig:naive-sets} (refer $k=1$ accuracy). Thus, our results demonstrate that the set size obtained from conformal prediction procedure can filter low-quality predictions in downstream applications for LLMs. For example, highly uncertain predictions in a disease screening application should be flagged for manual review and not shown to the user.
    
\textbf{Size-stratified coverage and comparison with naive top-$k$ prediction sets.}
    Size-stratified coverage measures error-rate guarantee across prediction sets of different sizes~\cite{angelopoulos2022uncertainty}.
    This experiment shows that coverage is not trivially satisfied by naively forming prediction sets by simply taking the top-$k$ highest softmax probabilities.
    In Figure~\ref{fig:naive-sets}, we show the coverage when all prediction sets have a fixed set size and find that coverage decreases sharply with size.
    This is in contrast to prediction sets formed by conformal prediction in Figure~\ref{fig:stratified-coverage}, where we find that even prediction sets of size one have close to the desired level of coverage ($90\%$ when $\alpha = 0.1$) across most subjects.
    Indeed, we found that coverage is consistent over all set sizes for conformal prediction.
    
    Conformal prediction can be thought of as outputting ``adaptive'' prediction sets that try to attain the proper level of coverage (depending on the chosen error rate $\alpha$) instead of ``fixed'' prediction sets of size $k$. 

\textbf{Exchangeability assumption across subjects.} \label{exchange}
    In Figure~\ref{fig:cross-calibration}, we test the exchangeability assumptions between subjects 
    by calibrating on one subject and evaluating coverage on a different subject, grouped into three categories of subjects. 
    Recall that the exchangeability assumption is needed for the coverage guarantee of Equation~\ref{eq:coverage-guarantee} to hold. 
    
    On the main diagonal, where the prediction sets are calibrated and evaluated on the same subject, we observed little deviation from the desired coverage rate of $90\%$.
    For example, prediction sets calibrated and evaluated on the same subject had close to the desired error rate of $10\%$ when $\alpha =0.1$.
    On the off-diagonal, we can see significant disparities between some subjects.
    For example, when prediction sets are calibrated on MCQA data from ``high school computer science'' and evaluated on ``business ethics'', coverage is only around $83\%$, less than the desired $90\%$ coverage.
    However, for subjects from similar domains and accuracy, such as ``clinical knowledge'', ``anatomy'', and ``high school biology'', we find relatively more minor deviations from the targeted coverage rate when calibrated on out-of-subject data. This may result from good generalization capabilities and relatively calibrated softmax probability~\cite{kadavath2022language} outputted by  the LLMs.

\subsection{Code Availability} \label{code}
    We release the code at this \href{https://github.com/bhaweshiitk/ConformalLLM/tree/main}{\textbf{\underline{Github repository}}}. The code repository also contains the question-answer pairs generated by GPT-4 for our prompts.

\section{Discussion} \label{summary}

As Large Language Models (LLMs) become increasingly powerful and are deployed in mission-critical systems, obtaining formal uncertainty guarantees for these models is crucial. 

In this work, we investigated uncertainty quantification in LLMs in the context of multiple-choice questions using conformal prediction, a statistical framework, for generating prediction sets with coverage guarantees. 

We found that naive softmax outputs of LLMs are relatively well calibrated on average but can suffer from under-confidence and over-confidence, and the extent of miscalibration varies across different subjects. To have a formal guarantee on the error rate of the model prediction, we implemented the conformal prediction procedure on the naive softmax output of the LLM. 

The conformal prediction framework produces valid prediction sets with error rate guarantees when calibration and evaluation sets come from the same distribution. 
We also explored the application of conformal prediction procedures for selective classification tasks. We found that conformal prediction can be used to discard predictions with unusual and low-quality outputs where the model is not confident, as indicated by the size of its prediction sets. 

To summarize, our main takeaways are
 \begin{itemize}
     \item Developers of LLM systems should provide estimates of uncertainty to improve trustworthiness in their outputs to users.

     \item Uncertainty quantification can be useful for downstream applications such as filtering biased, unusual, or low-quality outputs.
     
     \item Conformal prediction is one approach to uncertainty quantification where a user-specified error rate can be statistically guaranteed when the calibration data is exchangeable with the test data. 


     \item For our specific dataset (MMLU) and LLM (LLaMA-13B), we find that softmax outputs obtained as described in section~\ref{model_and_dataset} are reasonably calibrated on average. Nonetheless, models suffer from under-confidence and overconfidence, especially at the tail ends of probability distribution (refer figure ~\ref{fig:calibration-error} in the Appendix.) 
 \end{itemize}

Our work has some limitations. Our findings were limited to the MCQA task on the MMLU dataset using the LLaMA-13B model. Future works could extend our findings to multiple models and data sets. Further, it would be interesting to extend the conformal prediction framework to more general settings like free-form text generation to control for inaccurate, biased, and harmful outputs from LLMs. It would also be interesting to explore exchangeability conditions in LLMs further when calibration and evaluation data sets are from different distributions (i.e., not just from MMLU), which is a more realistic scenario. 

Despite these limitations, our work represents, to our knowledge, the first exploration of conformal prediction for LLMs in classification tasks. Our results contribute to the growing body of research on uncertainty estimation and generalization capabilities of LLMs and serve as a step forward in developing more robust and reliable uncertainty measures for increasingly capable large language models. Such measures are essential for ensuring LLMs' safe and responsible deployment in mission-critical applications.

\section*{Acknowledgement}  
We thank Prof. Yoon Kim,  Abbas Zeitoun, and Anastasios Angelopoulos for helpful discussions and feedback on this work.
\bibliography{example_paper}
\bibliographystyle{icml2023}

\newpage
\appendix
\onecolumn

\section{Appendix} \label{appendix}
\begin{table}[ht]
    \caption{Empirical coverage and prediction set size at two specified error rates.}
    \label{tab:results}
    \vskip 0.15in
    \begin{center}
    \begin{sc}
        \begin{tabular}{lc|c|c}
        Dataset & $1 - \alpha$ & Coverage & Set size \\
        \toprule
        \multirow{2}{*}{Professional accounting}
        & $90\%$ & $91\% \pm 3\%$ & $3.5 \pm 0.1$ \\
        & $80\%$ & $81\% \pm 3\%$ & $3.0 \pm 0.0$ \\
        \midrule
        \multirow{2}{*}{Business ethics}
        & $90\%$ & $93\% \pm 2\%$ & $3.4 \pm 0.1$ \\
        & $80\%$ & $82\% \pm 3\%$ & $2.8 \pm 0.2$ \\
        \midrule
        \multirow{2}{*}{Management}
        & $90\%$ & $94\% \pm 2\%$ & $3.1 \pm 0.1$ \\
        & $80\%$ & $83\% \pm 3\%$ & $2.5 \pm 0.1$ \\
        \midrule
        \multirow{2}{*}{Public relations}
        & $90\%$ & $93\% \pm 2\%$ & $3.0 \pm 0.1$ \\
        & $80\%$ & $83\% \pm 2\%$ & $2.3 \pm 0.1$ \\
        \midrule
        \multirow{2}{*}{Marketing}
        & $90\%$ & $91\% \pm 1\%$ & $2.4 \pm 0.1$ \\
        & $80\%$ & $81\% \pm 1\%$ & $1.6 \pm 0.1$ \\
        \midrule
        \multirow{2}{*}{College chemistry}
        & $90\%$ & $93\% \pm 2\%$ & $3.6 \pm 0.1$ \\
        & $80\%$ & $82\% \pm 4\%$ & $3.2 \pm 0.1$ \\
        \midrule
        \multirow{2}{*}{Professional medicine}
        & $90\%$ & $91\% \pm 6\%$ & $3.4 \pm 0.2$ \\
        & $80\%$ & $82\% \pm 7\%$ & $2.9 \pm 0.2$ \\
        \midrule
        \multirow{2}{*}{College medicine}
        & $90\%$ & $92\% \pm 2\%$ & $3.4 \pm 0.1$ \\
        & $80\%$ & $82\% \pm 2\%$ & $2.8 \pm 0.1$ \\
        \midrule
        \multirow{2}{*}{Clinical knowledge}
        & $90\%$ & $91\% \pm 3\%$ & $3.2 \pm 0.1$ \\
        & $80\%$ & $82\% \pm 3\%$ & $2.7 \pm 0.1$ \\
        \midrule
        \multirow{2}{*}{Anatomy}
        & $90\%$ & $92\% \pm 3\%$ & $3.3 \pm 0.1$ \\
        & $80\%$ & $81\% \pm 4\%$ & $2.7 \pm 0.1$ \\
        \midrule
        \multirow{2}{*}{High school biology}
        & $90\%$ & $91\% \pm 1\%$ & $3.2 \pm 0.1$ \\
        & $80\%$ & $81\% \pm 2\%$ & $2.6 \pm 0.1$ \\
        \midrule
        \multirow{2}{*}{Formal logic}
        & $90\%$ & $92\% \pm 2\%$ & $3.7 \pm 0.1$ \\
        & $80\%$ & $82\% \pm 3\%$ & $3.2 \pm 0.1$ \\
        \midrule
        \multirow{2}{*}{Machine learning}
        & $90\%$ & $93\% \pm 2\%$ & $3.6 \pm 0.1$ \\
        & $80\%$ & $82\% \pm 4\%$ & $3.1 \pm 0.1$ \\
        \midrule
        \multirow{2}{*}{College computer science}
        & $90\%$ & $93\% \pm 2\%$ & $3.5 \pm 0.2$ \\
        & $80\%$ & $83\% \pm 2\%$ & $3.1 \pm 0.2$ \\
        \midrule
        \multirow{2}{*}{High school computer science}
        & $90\%$ & $93\% \pm 2\%$ & $3.2 \pm 0.2$ \\
        & $80\%$ & $82\% \pm 3\%$ & $2.7 \pm 0.1$ \\
        \midrule
        \multirow{2}{*}{Computer Security}
        & $90\%$ & $94\% \pm 3\%$ & $2.9 \pm 0.1$ \\
        & $80\%$ & $83\% \pm 2\%$ & $2.2 \pm 0.1$ \\
        \bottomrule
        \end{tabular}
    \end{sc}
    \end{center}
    \vskip -0.1in
\end{table}

\begin{figure}
     \centering
     \begin{subfigure}[b]{0.24\textwidth}
         \centering
         \includegraphics[width=\textwidth]{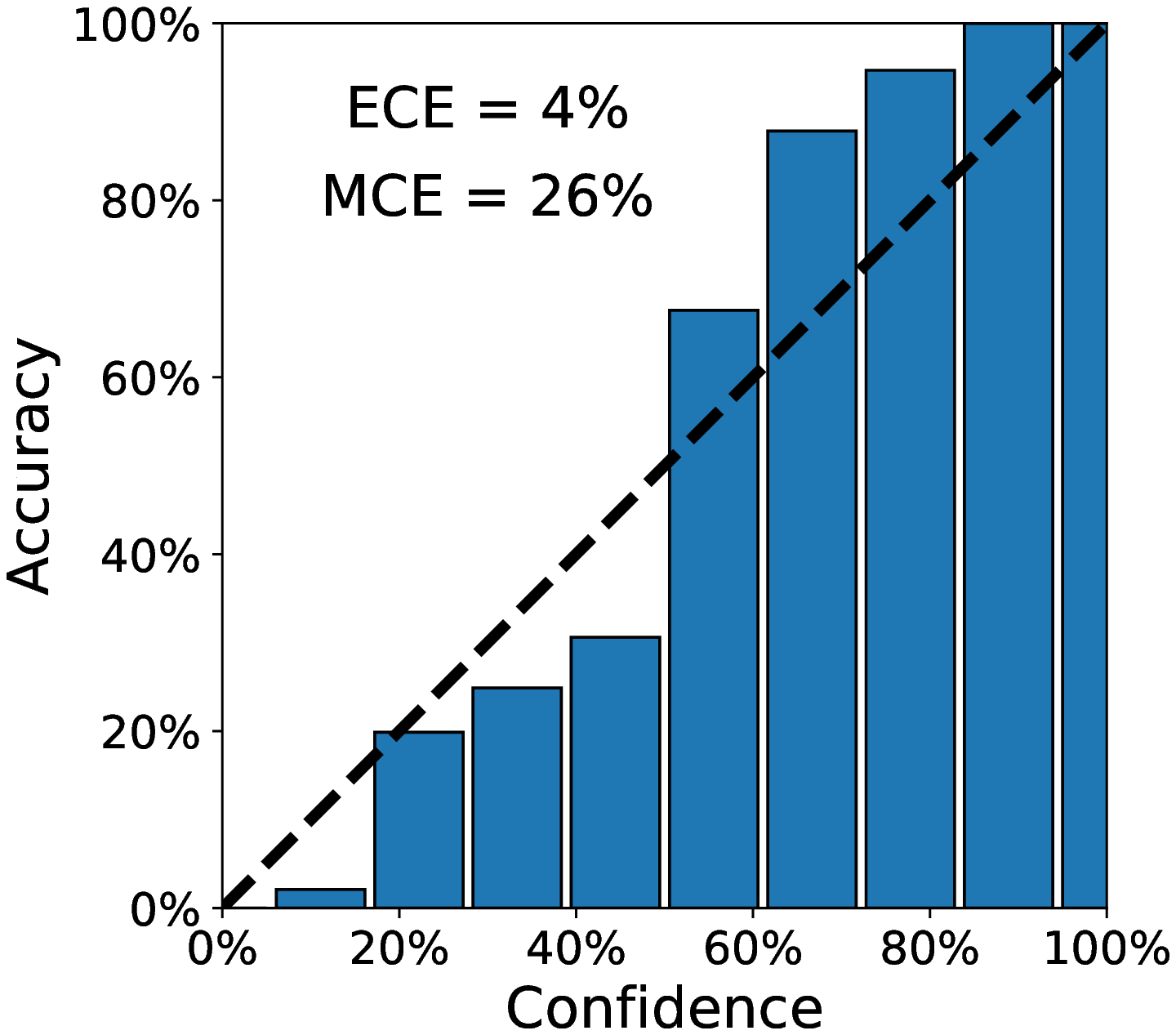}
         \caption{anatomy}
         \label{fig:anatomy}
     \end{subfigure}
     \begin{subfigure}[b]{0.24\textwidth}
         \centering
         \includegraphics[width=\textwidth]{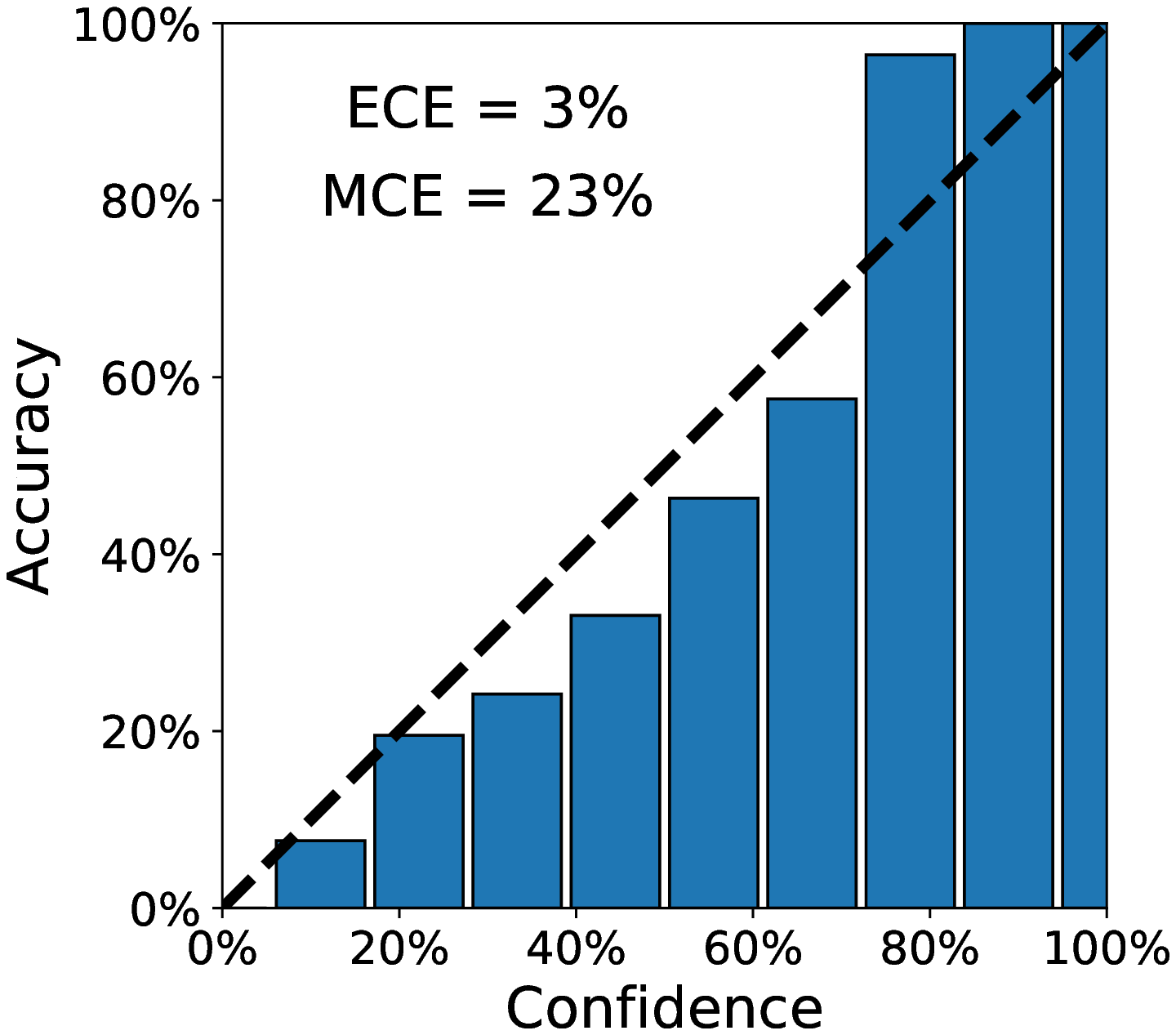}
         \caption{business ethics}
         \label{fig:business-ethics}
     \end{subfigure}
     \begin{subfigure}[b]{0.24\textwidth}
         \centering
         \includegraphics[width=\textwidth]{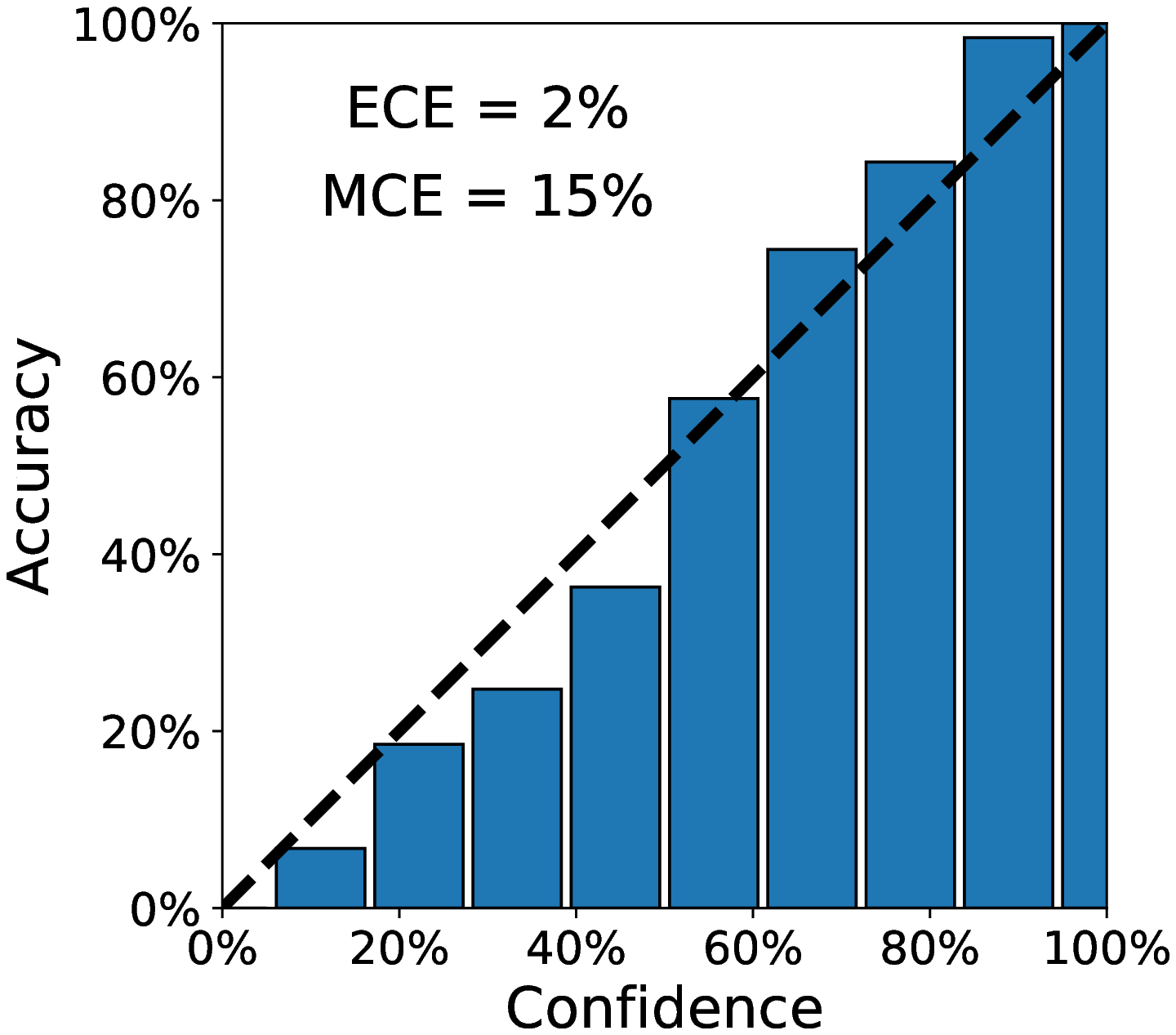}
         \caption{clinical knowledge}
         \label{fig:clinical-knowledge}
     \end{subfigure}
     \begin{subfigure}[b]{0.24\textwidth}
         \centering
         \includegraphics[width=\textwidth]{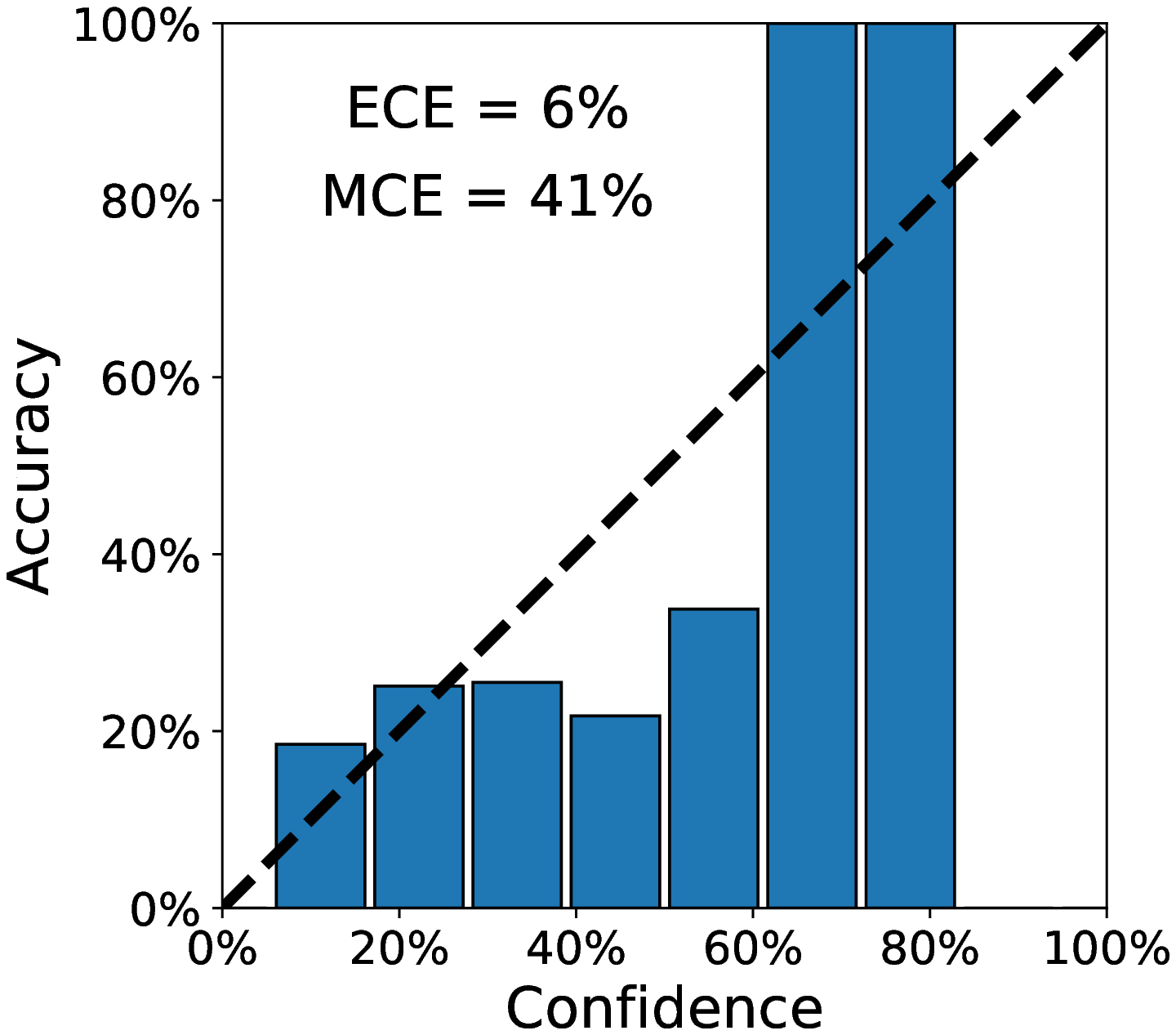}
         \caption{college chemistry}
         \label{fig:college-chemistry}
     \end{subfigure}
     \begin{subfigure}[b]{0.24\textwidth}
         \centering
         \includegraphics[width=\textwidth]{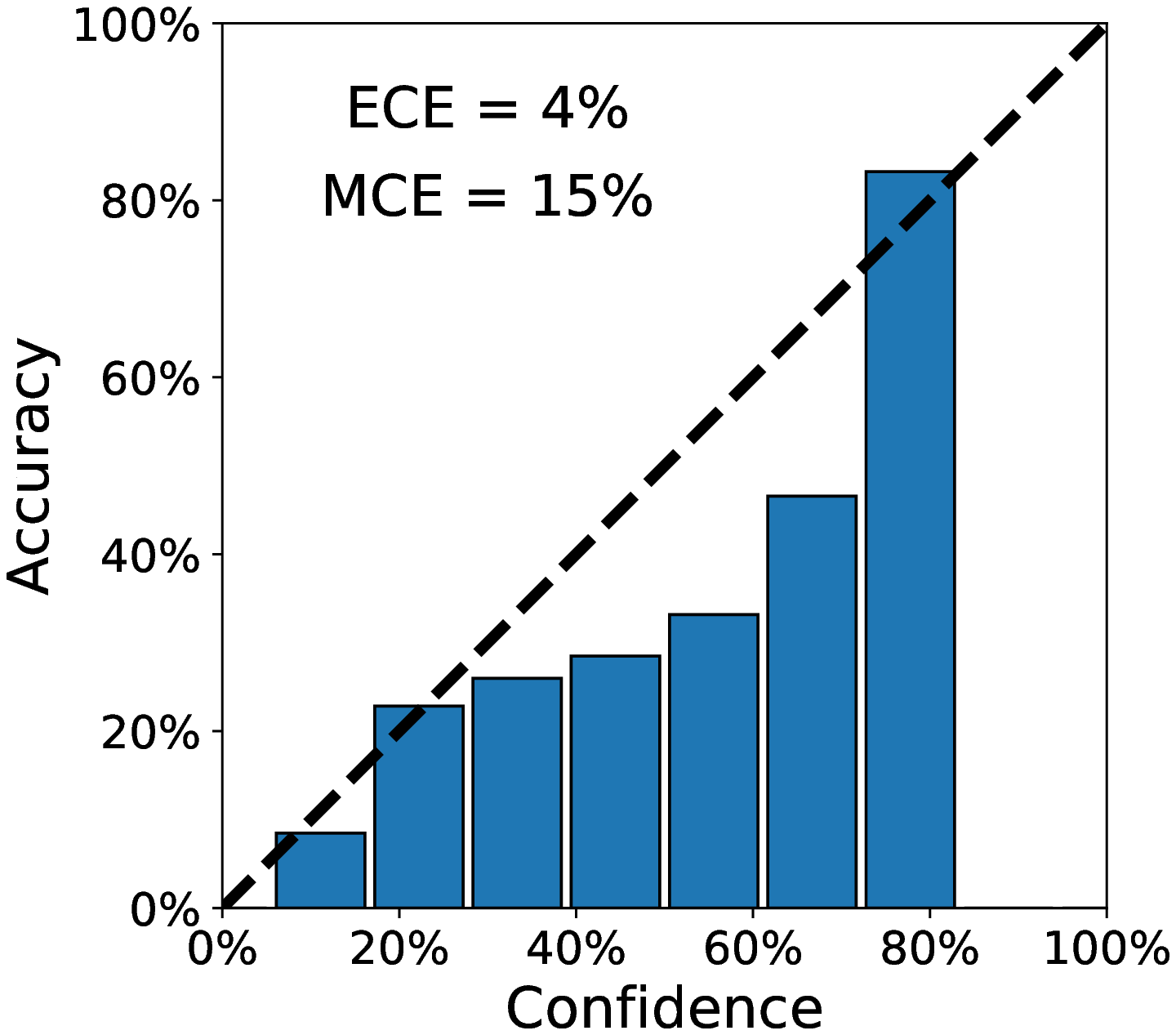}
         \caption{college computer science}
         \label{fig:college-cs}
     \end{subfigure}
     \begin{subfigure}[b]{0.24\textwidth}
         \centering
         \includegraphics[width=\textwidth]{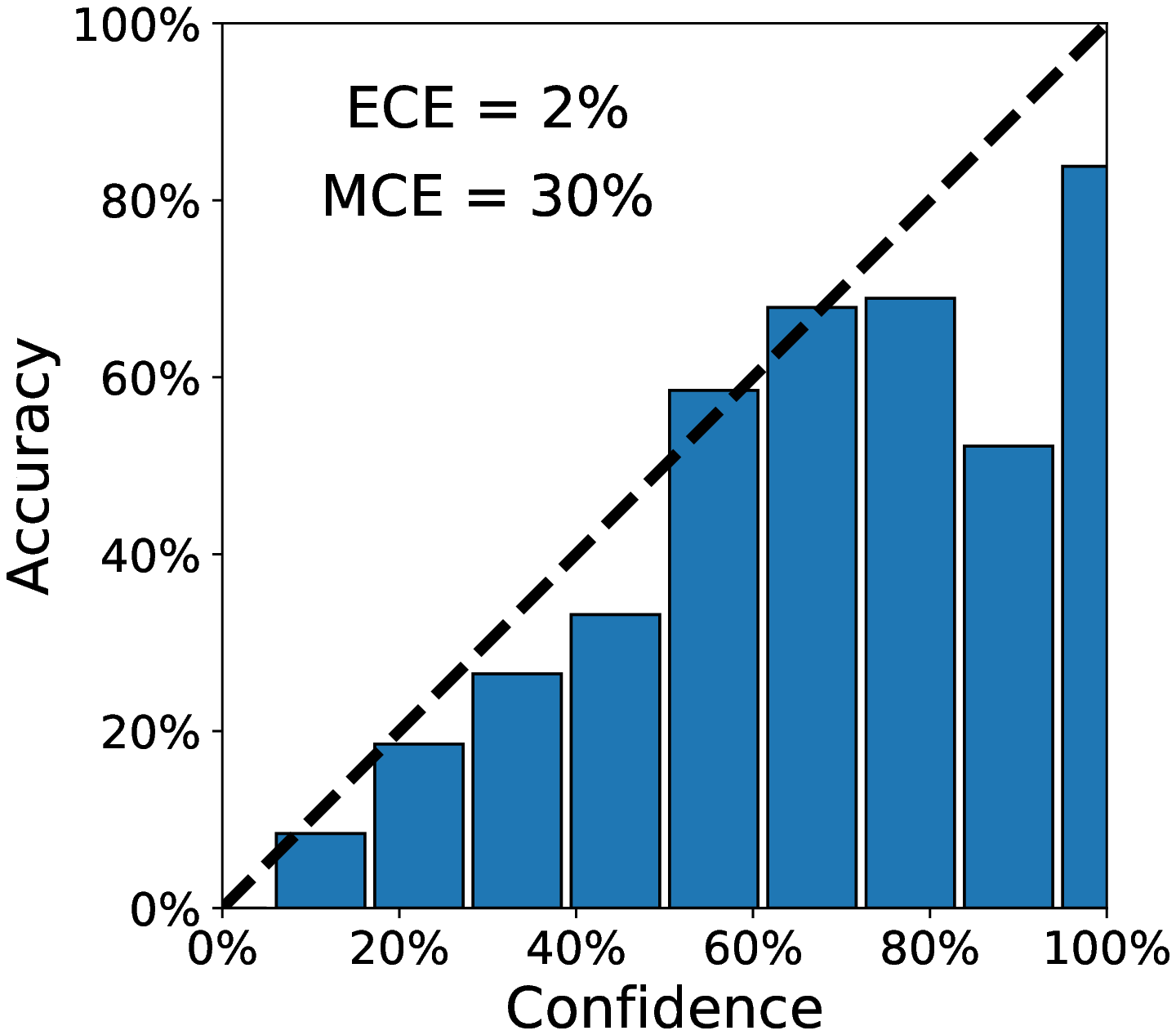}
         \caption{college medicine}
         \label{fig:college-medicine}
     \end{subfigure}
     \begin{subfigure}[b]{0.24\textwidth}
         \centering
         \includegraphics[width=\textwidth]{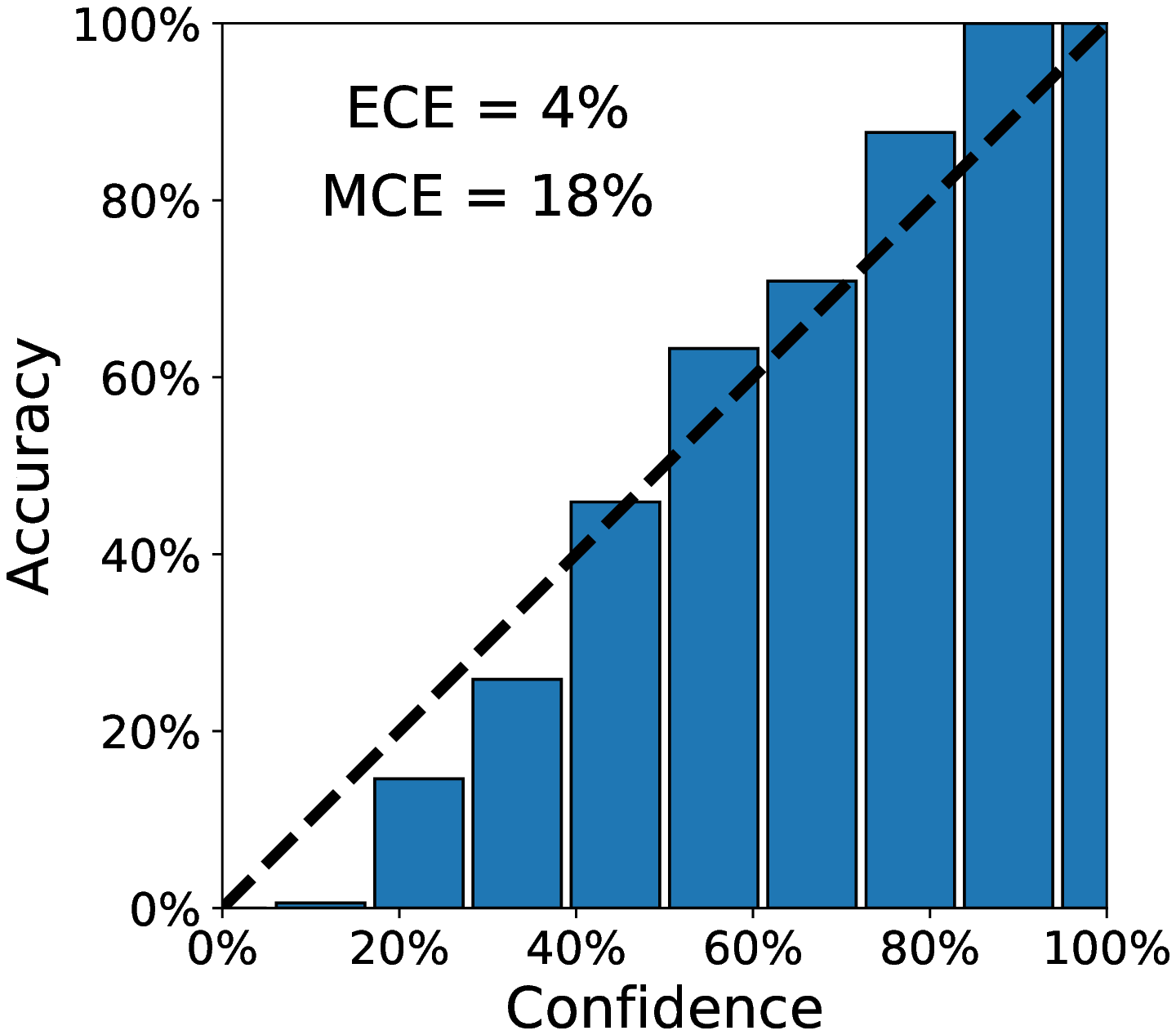}
         \caption{computer security}
         \label{fig:computer-security}
     \end{subfigure}
     \begin{subfigure}[b]{0.24\textwidth}
         \centering
         \includegraphics[width=\textwidth]{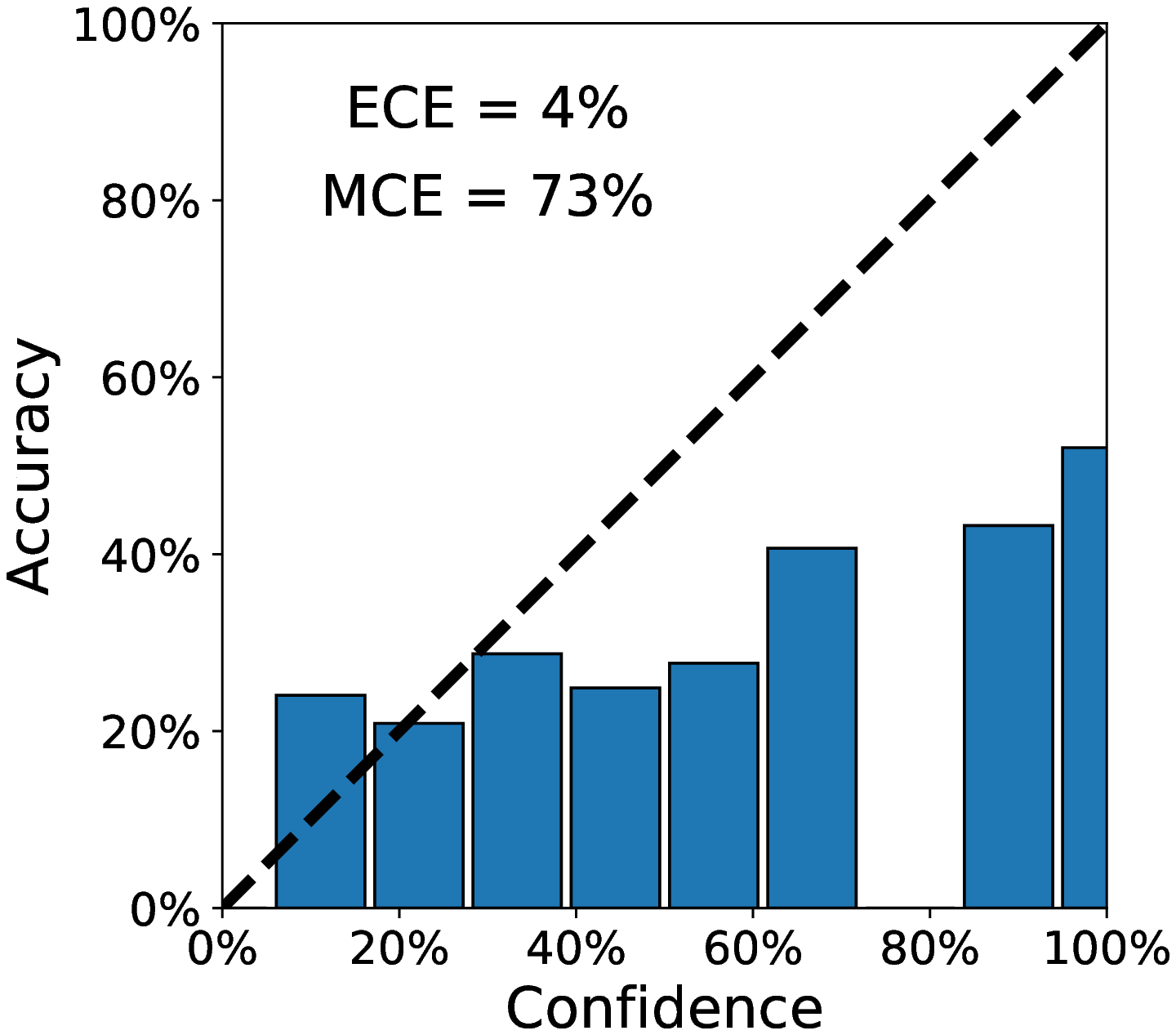}
         \caption{formal logic}
         \label{fig:formal-logic}
     \end{subfigure}
     \begin{subfigure}[b]{0.24\textwidth}
         \centering
         \includegraphics[width=\textwidth]{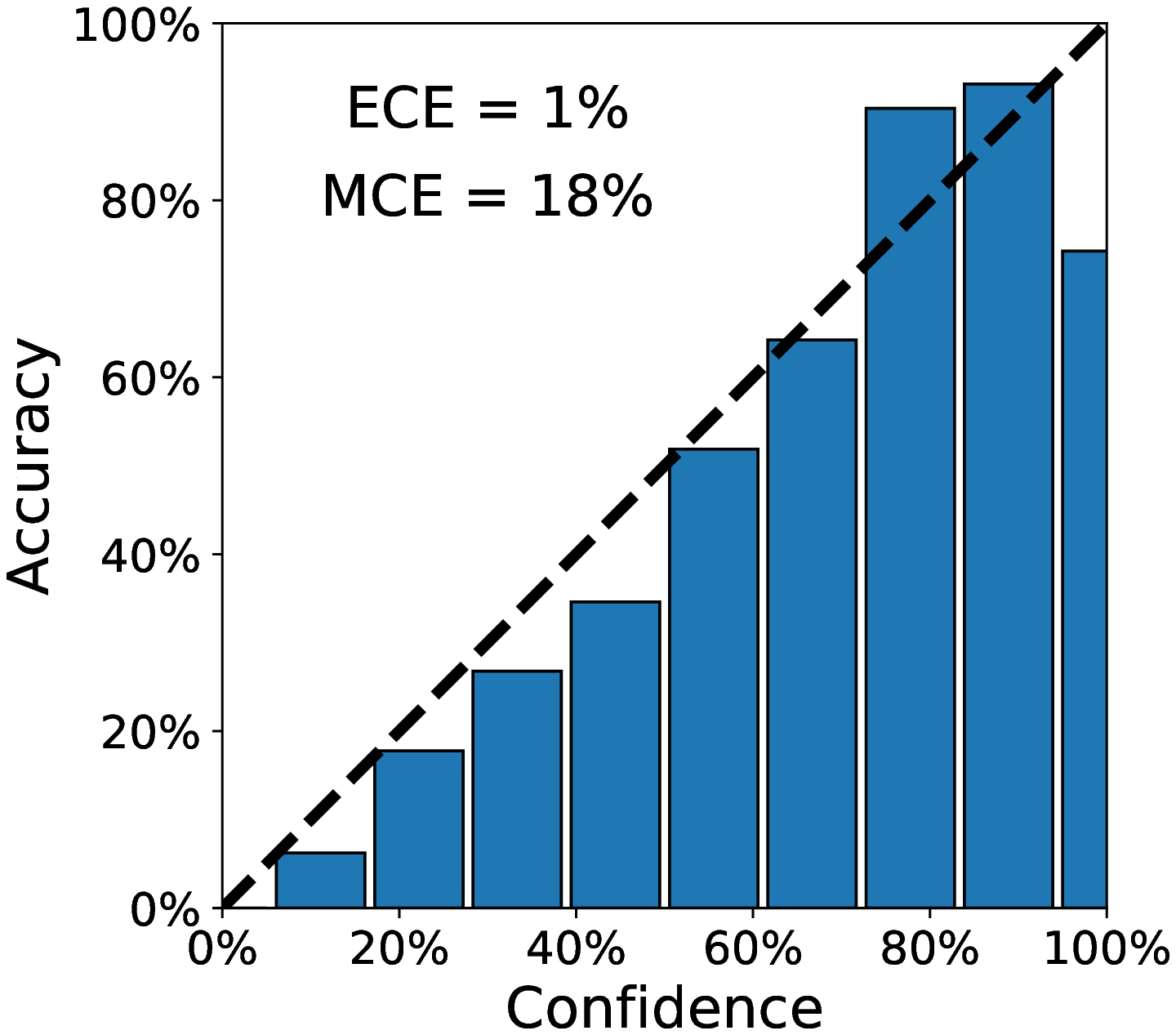}
         \caption{high school biology}
         \label{fig:high-school-biology}
     \end{subfigure}
     \begin{subfigure}[b]{0.24\textwidth}
         \centering
         \includegraphics[width=\textwidth]{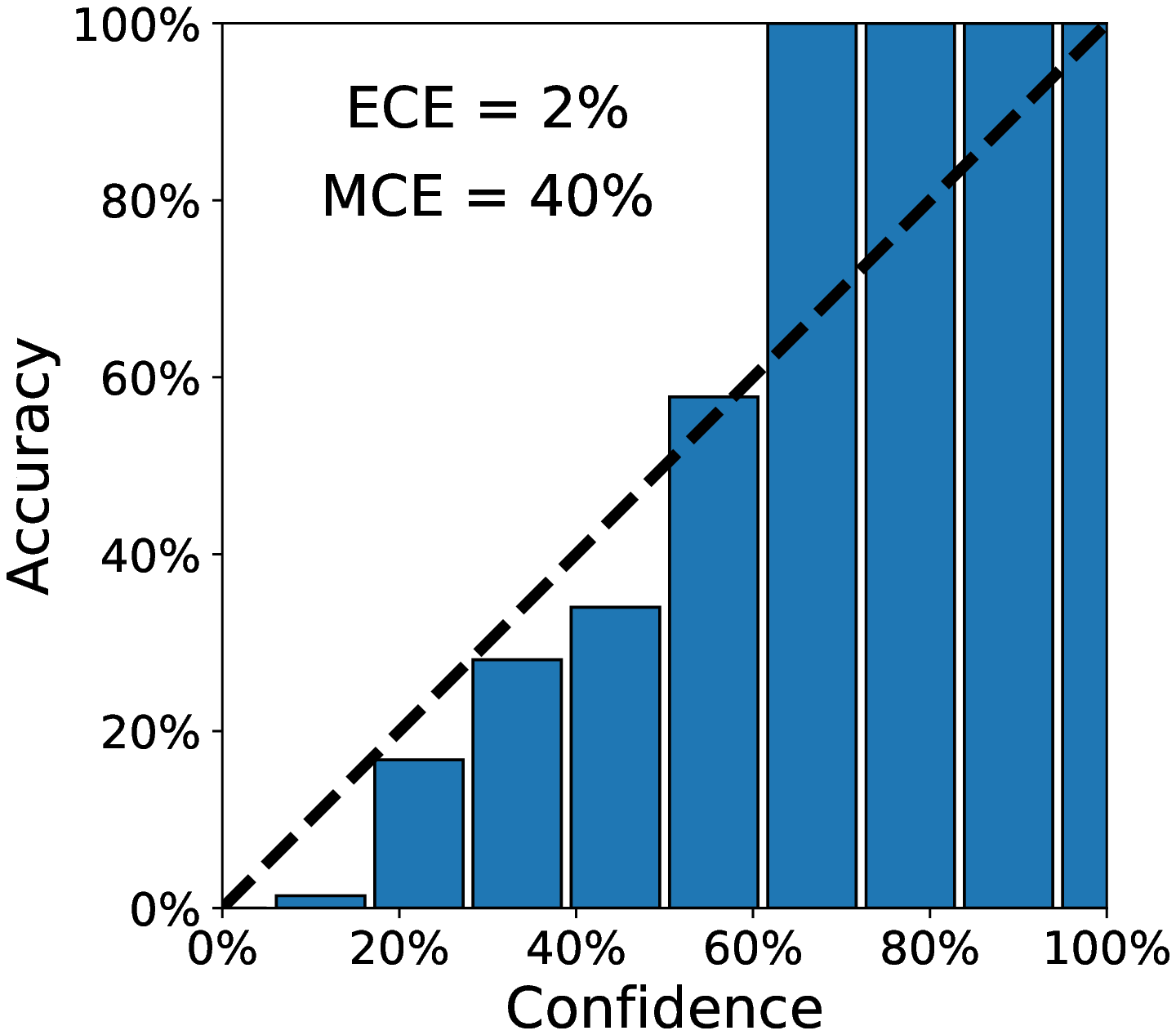}
         \caption{high school computer science}
         \label{fig:high-school-cs}
     \end{subfigure}
     \begin{subfigure}[b]{0.24\textwidth}
         \centering
         \includegraphics[width=\textwidth]{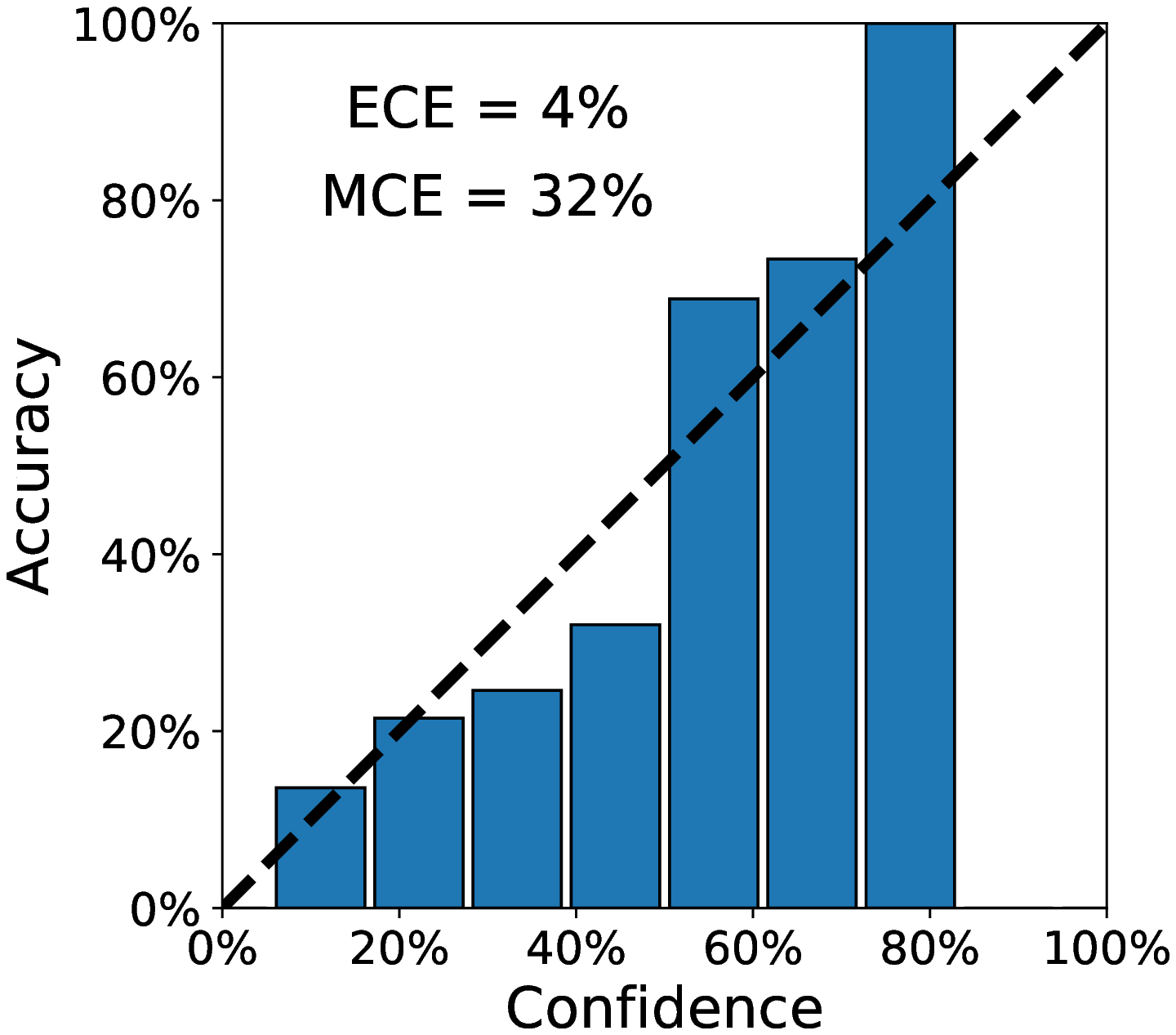}
         \caption{machine learning}
         \label{fig:machine-learning}
     \end{subfigure}
     \begin{subfigure}[b]{0.24\textwidth}
         \centering
         \includegraphics[width=\textwidth]{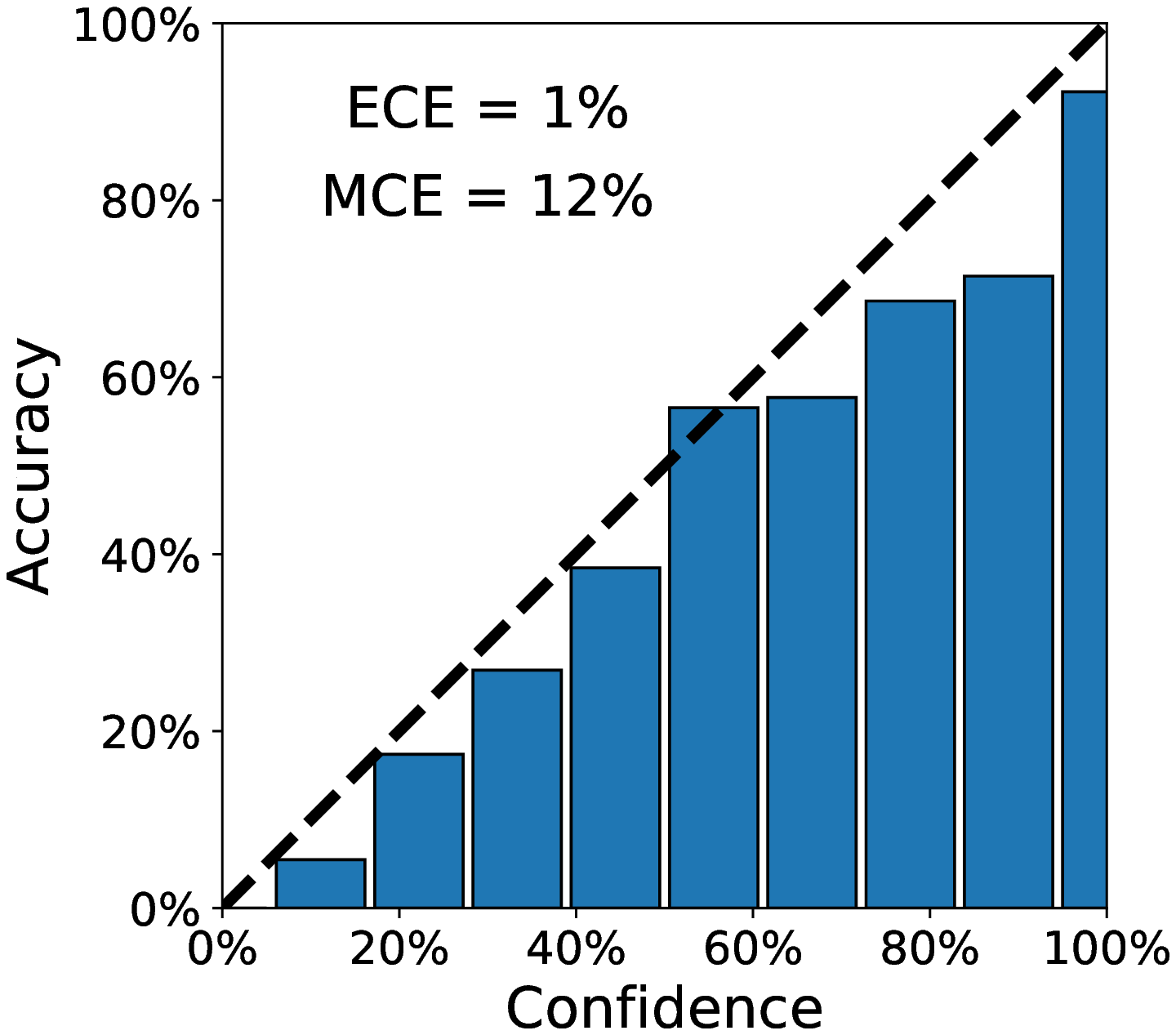}
         \caption{management}
         \label{fig:management}
     \end{subfigure}
     \begin{subfigure}[b]{0.24\textwidth}
         \centering
         \includegraphics[width=\textwidth]{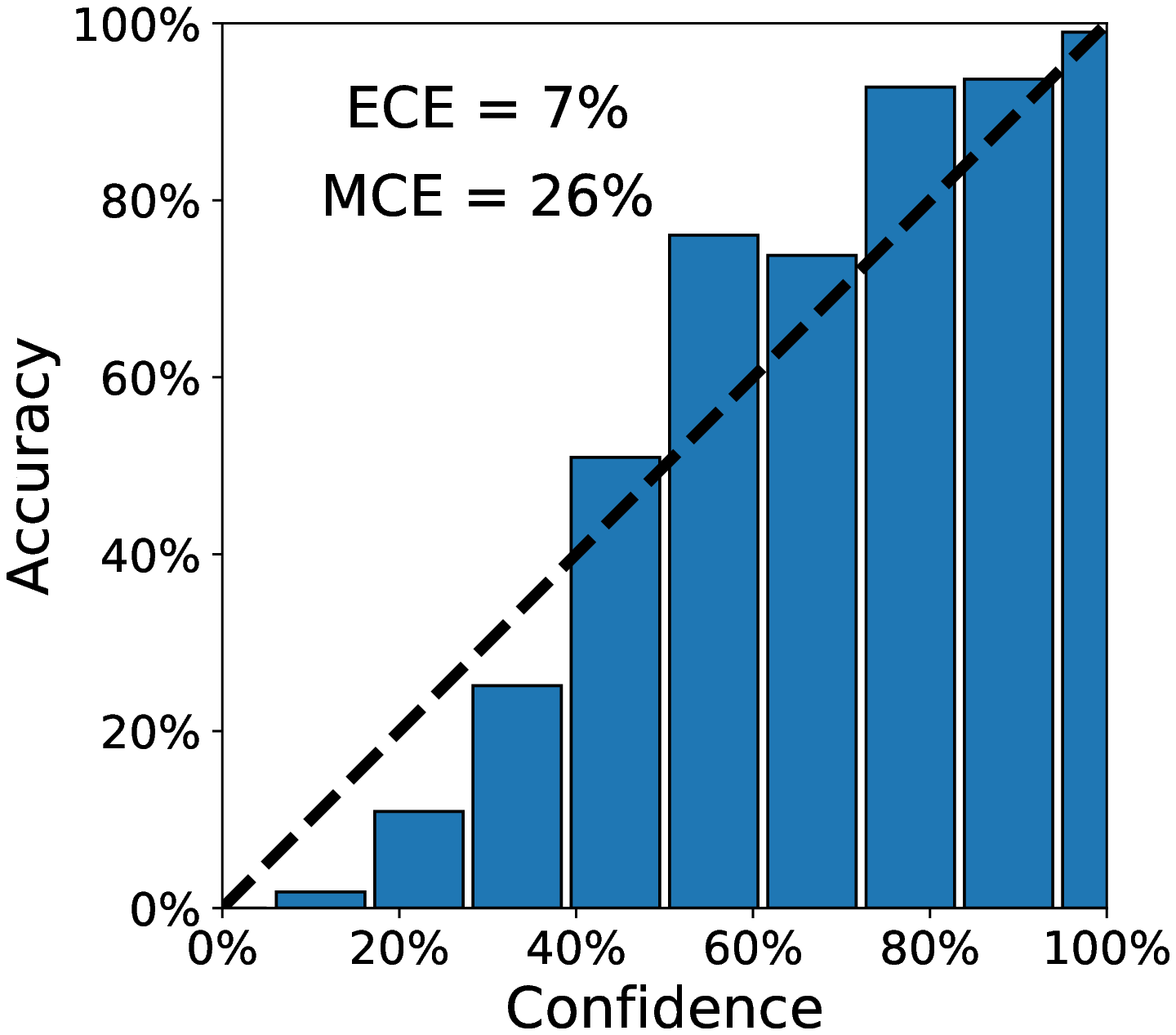}
         \caption{marketing}
         \label{fig:marketing}
     \end{subfigure}
     \begin{subfigure}[b]{0.24\textwidth}
         \centering
         \includegraphics[width=\textwidth]{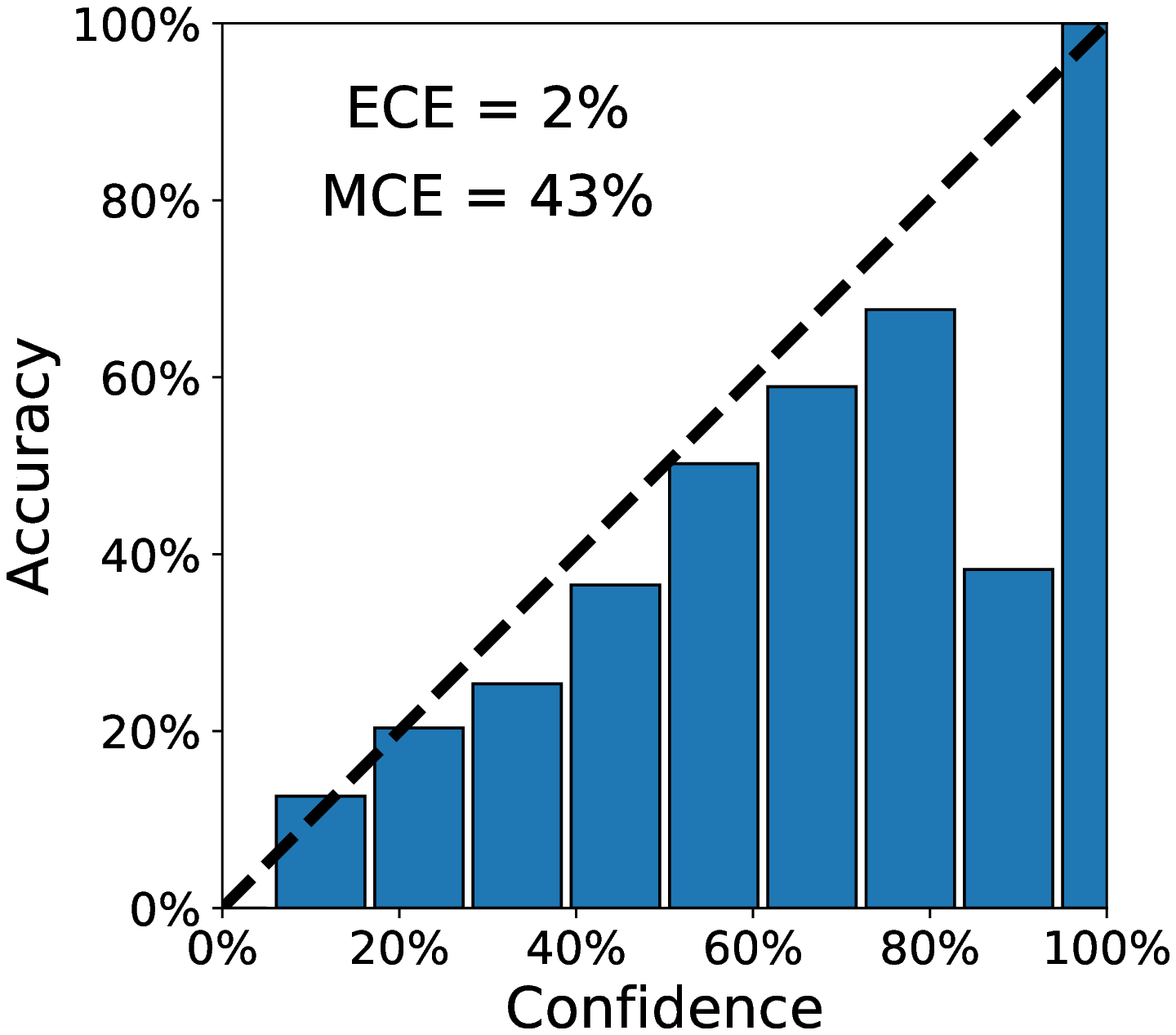}
         \caption{professional accounting}
         \label{fig:professional-accounting}
     \end{subfigure}
     \begin{subfigure}[b]{0.24\textwidth}
         \centering
         \includegraphics[width=\textwidth]{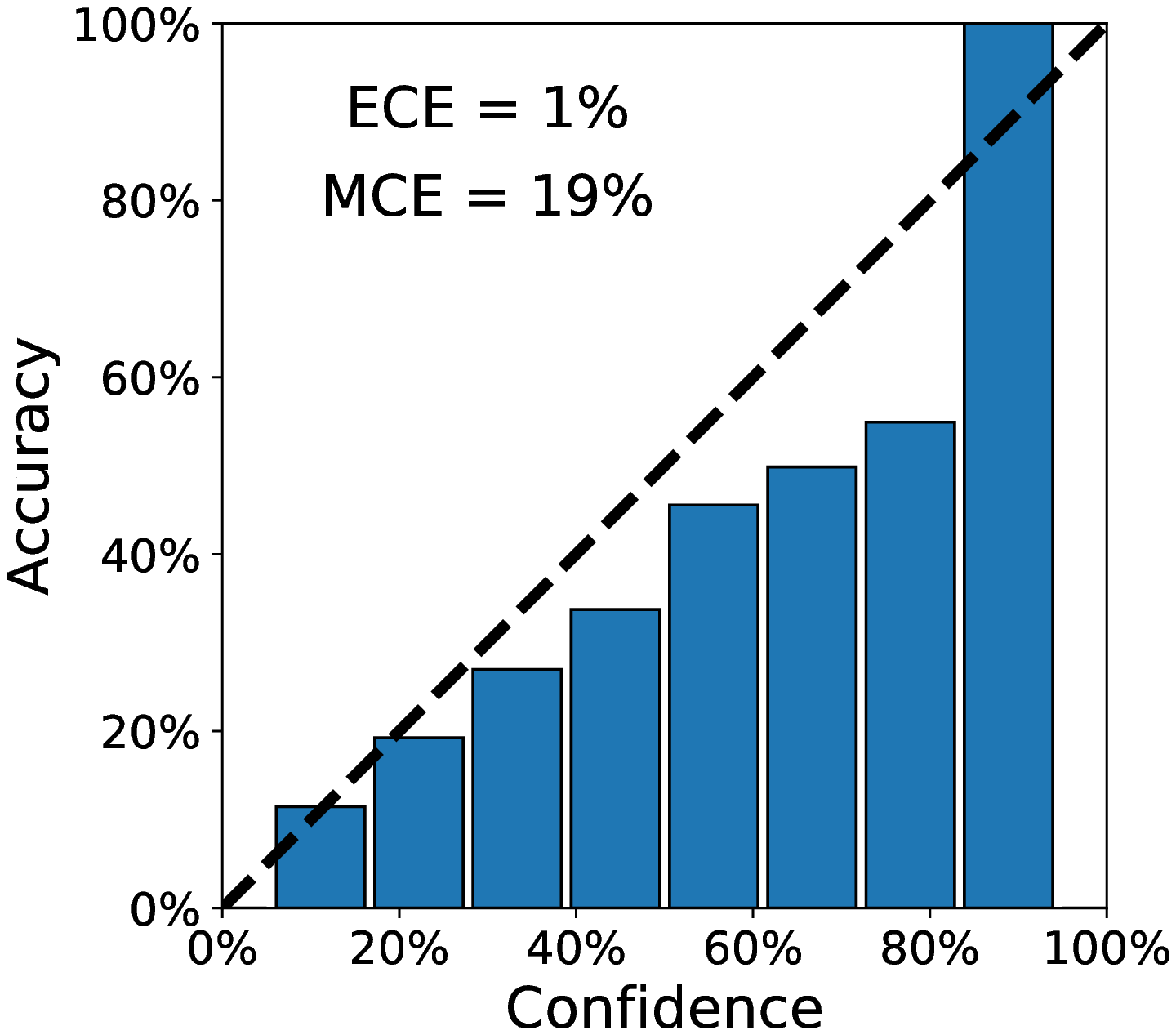}
         \caption{professional medicine}
         \label{fig:professional-medicine}
     \end{subfigure}
     \begin{subfigure}[b]{0.24\textwidth}
         \centering
         \includegraphics[width=\textwidth]{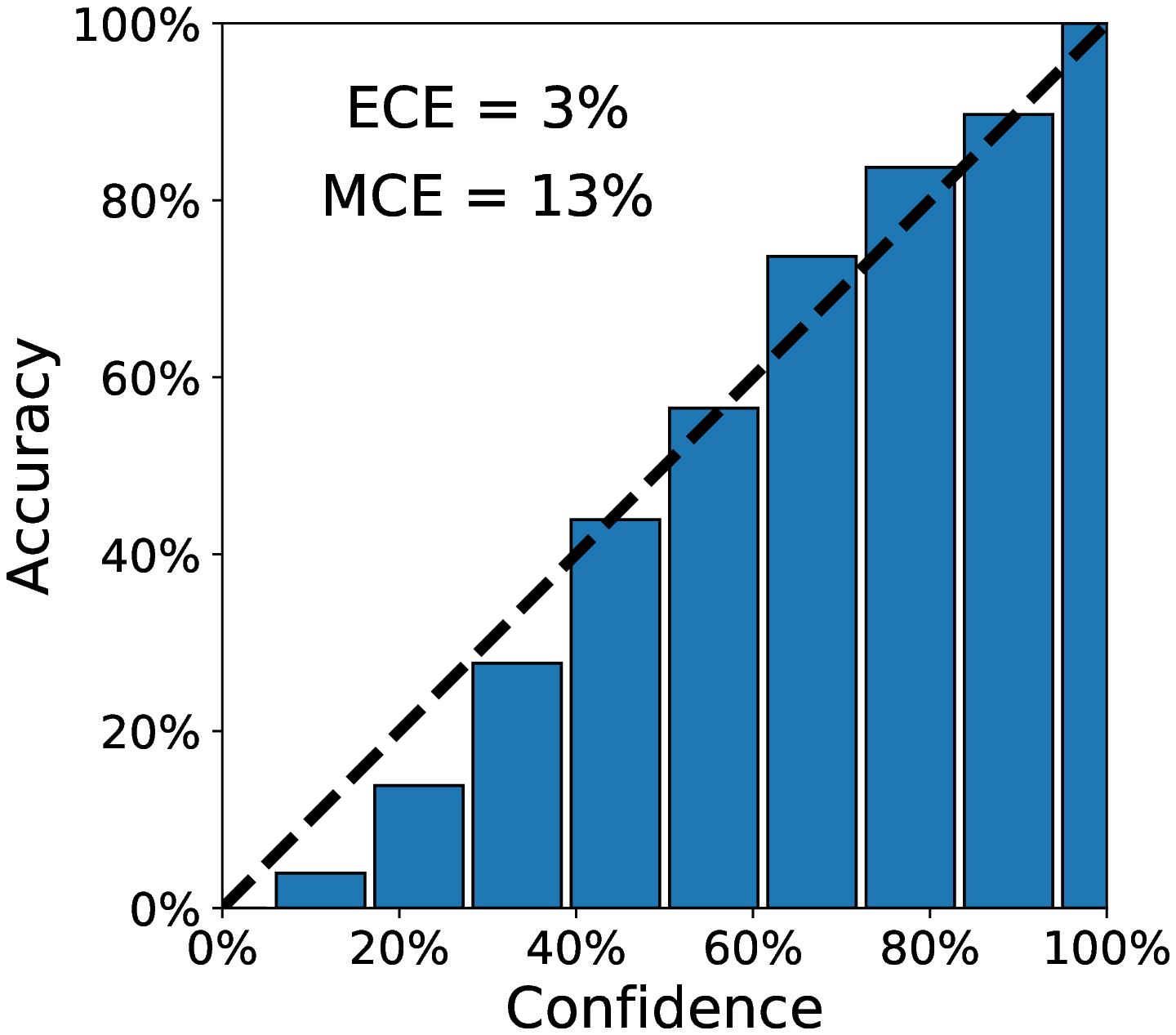}
         \caption{public relations}
         \label{fig:public relations}
     \end{subfigure}
     
        \caption{\textbf{Maximum softmax confidence does not represent true probability.} Deviation of softmax confidence from the probability of being correct for each subject. ECE is the expected calibration error, and MCE is the maximum calibration error.}
        \label{fig:calibration-error}
\end{figure}
\clearpage
\subsection{Sample GPT-4- and MMLU-based one-shot questions for selected subjects.}\label{appendix_qa}

\subsubsection{College Computer Science}
GPT-4 Based One-shot Questions
\begin{verbatim}
Which of the following sorting algorithms has the best average case performance?
    A. Bubble Sort
    B. Quick Sort
    C. Selection Sort
    D. Insertion Sort
The correct answer is option: B

What does the term "Big O Notation" describe in Computer Science?
    A. The speed of a computer
    B. The operating system version
    C. The size of a database
    D. The time complexity of an algorithm
The correct answer is option: D

What does HTTP stand for in terms of web technology?
    A. Hyper Text Transfer Portal
    B. Hyper Transfer Protocol
    C. Hyper Text Transfer Protocol
    D. High Transfer Text Protocol
The correct answer is option: C

In object-oriented programming, what is 'inheritance' used for?
    A. To distribute data across multiple databases
    B. To share methods and fields between classes
    C. To encrypt data before storing it
    D. To speed up program execution
The correct answer is option: B

Which of the following data structures is non-linear?
    A. Array
    B. Stack
    C. Tree
    D. Queue
The correct answer is option: C
\end{verbatim}
MMLU Based One-shot Questions
\begin{verbatim}
An integer c is a common divisor of two integers x and y if
and only if c is a divisor of x and c is a divisor of y. 
Which of the following sets of integers could possibly be
the set of all common divisors of two integers?
    A. {-6,-2, -1, 1, 2, 6}
    B. {-6, -2, -1, 0, 1, 2, 6}
    C. {-6, -3, -2, -1, 1, 2, 3, 6}
    D. {-6, -3, -2, -1, 0, 1, 2, 3, 6}
The correct answer is option: C.

You want to cluster 7 points into 3 clusters using the k-Means
Clustering algorithm. Suppose after the first iteration, clusters
C1, C2 and C3 contain the following two-dimensional points: C1
contains the 2 points: {(0,6), (6,0)} C2 contains the 3 points:
{(2,2), (4,4), (6,6)} C3 contains the 2 points: {(5,5), (7,7)}
What are the cluster centers computed for these 3 clusters?
    A. C1: (3,3), C2: (4,4), C3: (6,6)
    B. C1: (3,3), C2: (6,6), C3: (12,12)
    C. C1: (6,6), C2: (12,12), C3: (12,12)
    D. C1: (0,0), C2: (48,48), C3: (35,35)
The correct answer is option: A.

Consider the collection of all undirected graphs with 10 nodes
and 6 edges. Let M and m, respectively, be the maximum and
minimum number of connected components in any graph in the
collection. If a graph has no selfloops and there is at most
one edge between any pair of nodes, which of the following is true?
    A. M = 10, m = 10
    B. M = 10, m = 1
    C. M = 7, m = 4
    D. M = 6, m = 4
The correct answer is option: C.

Which of the following statements describe(s) properties of
a purely segmented memory system?
I. It divides memory into units of equal size.
II. It permits implementation of virtual memory.
III. It suffers from internal fragmentation.
    A. I only
    B. II only
    C. III only
    D. I and III
The correct answer is option: B.

Which of the following statements about floating-point arithmetic is NOT true?
    A. It is inherently nonassociative because some numbers have no      exact representation.
    B. It is inherently nonassociative because there have to be upper
       and lower bounds on the size of numbers.
    C. Associativity can be achieved with appropriate roundoff      
       conventions.
    D. Some rational numbers have no exact representation.
The correct answer is option: C.
\end{verbatim}

\subsubsection{Professional Accounting}
GPT-4 Based One-shot Questions
\begin{verbatim}
Which of the following is used in accounting to analyze the
financial health of a business?
    A. Horizontal analysis
    B. Vertical analysis
    C. Ratio analysis
    D. All of the above
The correct answer is option: D

What does the acronym GAAP stand for in accounting?
    A. General Accepted Accounting Principles
    B. Global Accepted Accounting Procedures
    C. General Applied Accounting Procedures
    D. Global Applied Accounting Principles
The correct answer is option: A

What is the basic accounting equation?
    A. Assets = Liabilities + Owner’s Equity
    B. Assets = Liabilities - Owner’s Equity
    C. Assets + Liabilities = Owner’s Equity
    D. Assets - Liabilities = Owner’s Equity
The correct answer is option: A

What is a balance sheet used for in accounting?
    A. To record the day-to-day financial transactions
    B. To determine the company's financial position at a specific point in time
    C. To track the company's cash flows
    D. To record the company's sales revenue
The correct answer is option: B

Which of the following best describes accrual accounting?
    A. Revenue and expenses are recorded when they are received and paid
    B. Revenue and expenses are recorded when they are earned and incurred
    C. Revenue and expenses are recorded at the end of the financial year
    D. Revenue and expenses are recorded at the start of the financial year
The correct answer is option: B

\end{verbatim}
MMLU Based One-shot Questions
\begin{verbatim}
Arno Co. did not record a credit purchase of merchandise made
prior to year end. However the merchandise was correctly included
in the year-end physical inventory. What effect did the omission
of reporting the purchase of merchandise have on Arno's balance
sheet at year end? Assets Liabilities
    A. No effect No effect
    B. No effect Understated
    C. Understated No effect
    D. Understated Understated
The correct answer is option B.

Which of the following procedures would an auditor generally perform
regarding subsequent events?
    A. Inspect inventory items that were ordered before the year end
       but arrived after the year end.
    B. Test internal control activities that were previously reported
       to management as inadequate.
    C. Review the client's cutoff bank statements for several months
       after the year end.
    D. Compare the latest available interim financial statements with
       the statements being audited.
The correct answer is option D.

The FASB makes changes to the Accounting Standards Codification by issuing
    A. Accounting Standards Updates.
    B. Emerging Issues Task Force Releases.
    C. Statements of Financial Accounting Standards.
    D. Staff Technical Bulletins.
The correct answer is option A.

On July 1 year 7 Dean Co. issued at a premium bonds with a due date
of July 1 year 12. Dean incorrectly used the straight-line method
instead of the effective interest method to amortize the premium.
How were the following amounts affected by the error at June 30
year 12? Bond carrying Retained amount earnings
    A. Overstated Understated
    B. Understated Overstated
    C. Overstated No effect
    D. No effect No effect
The correct answer is option D.

A company recently moved to a new building. The old building is
being actively marketed for sale, and the company expects to
complete the sale in four months. Each of the following statements
is correct regarding the old building, except:
    A. It will be reclassified as an asset held for sale.
    B. It will be classified as a current asset.
    C. It will no longer be depreciated.
    D. It will be valued at historical cost.
The correct answer is option D.
\end{verbatim}

\subsubsection{Clinical knowledge}
GPT-4 Based One-shot Questions
\begin{verbatim}
Which of the following is the most common cause of community-acquired pneumonia?
    A. Streptococcus pneumoniae
    B. Haemophilus influenzae
    C. Klebsiella pneumoniae
    D. Pseudomonas aeruginosa
The correct answer is option: A

Which hormone is primarily responsible for regulating blood calcium levels?
    A. Calcitonin
    B. Parathyroid hormone
    C. Thyroxine
    D. Insulin
The correct answer is option: B
    
What is the most common cause of acute pancreatitis?
    A. Gallstones
    B. Alcohol
    C. Hypertriglyceridemia
    D. Medications
The correct answer is option: A

What is the most common cause of secondary hypertension?
    A. Renal artery stenosis
    B. Pheochromocytoma
    C. Hyperaldosteronism
    D. Cushing's syndrome
The correct answer is option: A
   
Which of the following is a common extraintestinal manifestation
of ulcerative colitis?
    A. Erythema nodosum
    B. Gallstones
    C. Uveitis
    D. All of the above
The correct answer is option: D
\end{verbatim}

MMLU Based One-shot Questions
\begin{verbatim}
The key attribute in successful marathon running is:
    A. strength.
    B. power.
    C. stride length.
    D. stamina.
The correct answer is option D.

Which of the following is NOT a symptom of anaphylaxis?
    A. Stridor.
    B. Bradycardia.
    C. Severe wheeze.
    D. Rash.
The correct answer is option B.

In what situation are closed pouches applied?
    A. The patient has a semi-formed or liquid output.
    B. The patient has a colostomy.
    C. In the immediate post-operative period.
    D. The patient has a urostomy.
The correct answer is option B.

With an increasing number of sprints the:
    A. anaerobic contribution progressively increases.
    B. pH of the muscle falls below 6.0.
    C. blood glucose concentration falls below 3 mmol/L.
    D. relative contribution of aerobic metabolism increases.
The correct answer is option D.

Which of the following is true in diplopia?
    A. Diplopia can never occur if one eye is covered
    B. The outer image is always the false image
    C. A fourth nerve palsy occurs when the patient looks upwards
    D. A sixth nerve palsy causes a divergent squint
The correct answer is option B.

\end{verbatim}



\end{document}